**MANUSCRIPT**

# GENKL: An Iterative Framework for Resolving Label Ambiguity and Label Non-conformity in Web Images Via a New GENeralized KL Divergence

**Xia Huang[1] · Kai Fong Ernest Chong[1]**

**Abstract**

Web image datasets curated online inherently contain ambiguous in-distribution instances and out-of-distribution instances, which we collectively call *non-conforming* (NC) instances. In many recent approaches for mitigating the negative effects of NC instances, the core implicit assumption is that the NC instances can be found via entropy maximization. For "entropy" to be well-defined, we are interpreting the output prediction vector of an instance as the parameter vector of a multinomial random variable, with respect to some trained model with a softmax output layer. Hence, entropy maximization is based on the idealized assumption that NC instances have predictions that are "almost" uniformly distributed. However, in real-world web image datasets, there are numerous NC instances whose predictions are far from being uniformly distributed. To tackle the limitation of entropy maximization, we propose $(\alpha, \beta)$-generalized KL divergence, $\mathcal{D}_{KL}^{\alpha,\beta}(p\|q)$, which can be used to identify significantly more NC instances. Theoretical properties of $\mathcal{D}_{KL}^{\alpha,\beta}(p\|q)$ are proven, and we also show empirically that a simple use of $\mathcal{D}_{KL}^{\alpha,\beta}(p\|q)$ outperforms all baselines on the NC instance identification task. Building upon $(\alpha, \beta)$-generalized KL divergence, we also introduce a new iterative training framework, GENKL, that identifies and relabels NC instances. When evaluated on three web image datasets, Clothing1M, Food101/Food101N, and mini WebVision 1.0, we achieved new state-of-the-art classification accuracies: 81.34%, 85.73% and 78.99%/92.54% (top-1/top-5), respectively.

## 1 Introduction

Web data is an abundant source for curating image datasets (Bossard et al., 2014; Kaur et al., 2017; Lee et al., 2018; Liang et al., 2020; Shang et al., 2018; Xiao et al., 2015). Raw web images collected online are typically annotated with weak-supervision methods (Xiao et al. 2015; Varma & Ré 2018; Tekumalla & Banda 2021; Zhang et al. 2021; Helmstetter & Paulheim 2021; Yang et al. 2022). Although much more efficient in comparison to manual annotation, such automated annotation methods inevitably introduce *non-conforming* (NC) instances, which comprise both ambiguous in-distribution (ID) instances and out-of-distribution (OOD)

✉ Kai Fong Ernest Chong
  ernest_chong@sutd.edu.sg

  Xia Huang
  xia_huang@mymail.sutd.edu.sg

[1] Singapore University of Technology and Design, 8 Somapah Road, Singapore 487372, Singapore

instances. For example, Clothing1M (Xiao et al., 2015), a large-scale web image dataset that is well-known for containing real-world label noise, also contains NC instances; see Fig. 1 for explicit examples. Whether ambiguous ID or OOD, these NC instances may lead to significant performance degradation during training, which is not surprising since neural networks are capable of achieving zero training error even in the extreme case of completely noisy data (Arpit et al., 2017; Yao et al., 2021; Zhang et al., 2021). How then do we deal with such NC instances in web image datasets?

Numerous works (Goldberger & Ben-Reuven, 2016; Hendrycks et al., 2018; Ma et al., 2020; Patrini et al., 2017; Peng et al., 2020; Sharma et al., 2020; Xia et al., 2019; Yao et al., 2020) have tackled this problem by viewing NC instances from the lens of label noise. The underlying assumption is that NC instances hurt performance because they (may) have incorrect labels. From this viewpoint, the problem is then reduced to a simpler one: How do we alleviate the effect of label noise? However, such a simplification ignores the

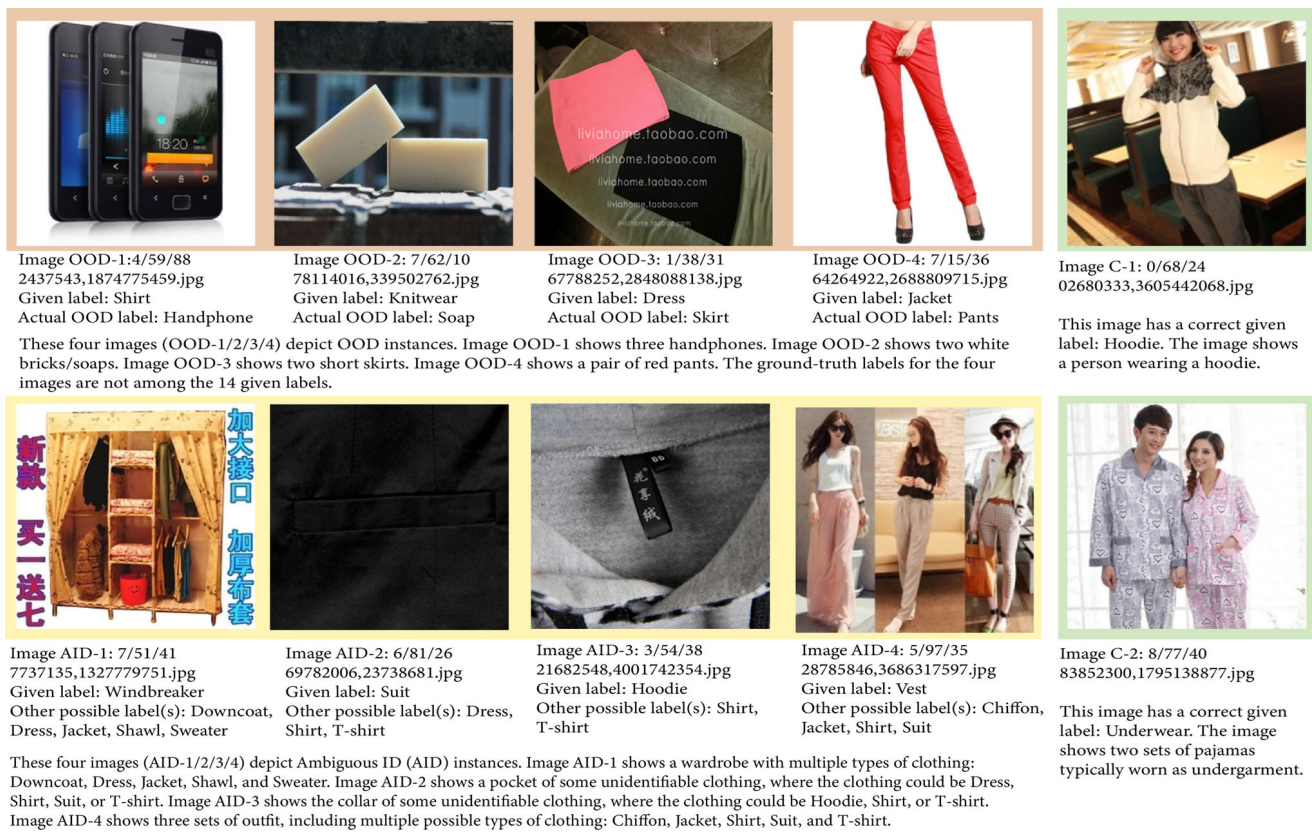

Image OOD-1:4/59/88
2437543,1874775459.jpg
Given label: Shirt
Actual OOD label: Handphone

Image OOD-2: 7/62/10
78114016,339502762.jpg
Given label: Knitwear
Actual OOD label: Soap

Image OOD-3: 1/38/31
67788252,2848088138.jpg
Given label: Dress
Actual OOD label: Skirt

Image OOD-4: 7/15/36
64264922,2688809715.jpg
Given label: Jacket
Actual OOD label: Pants

Image C-1: 0/68/24
02680333,3605442068.jpg

This image has a correct given label: Hoodie. The image shows a person wearing a hoodie.

These four images (OOD-1/2/3/4) depict OOD instances. Image OOD-1 shows three handphones. Image OOD-2 shows two white bricks/soaps. Image OOD-3 shows two short skirts. Image OOD-4 shows a pair of red pants. The ground-truth labels for the four images are not among the 14 given labels.

Image AID-1: 7/51/41
7737135,1327779751.jpg
Given label: Windbreaker
Other possible label(s): Downcoat, Dress, Jacket, Shawl, Sweater

Image AID-2: 6/81/26
69782006,23738681.jpg
Given label: Suit
Other possible label(s): Dress, Shirt, T-shirt

Image AID-3: 3/54/38
21682548,4001742354.jpg
Given label: Hoodie
Other possible label(s): Shirt, T-shirt

Image AID-4: 5/97/35
28785846,3686317597.jpg
Given label: Vest
Other possible label(s): Chiffon, Jacket, Shirt, Suit

Image C-2: 8/77/40
83852300,1795138877.jpg

This image has a correct given label: Underwear. The image shows two sets of pajamas typically worn as undergarment.

These four images (AID-1/2/3/4) depict Ambiguous ID (AID) instances. Image AID-1 shows a wardrobe with multiple types of clothing: Downcoat, Dress, Jacket, Shawl, and Sweater. Image AID-2 shows a pocket of some unidentified clothing, where the clothing could be Dress, Shirt, Suit, or T-shirt. Image AID-3 shows the collar of some unidentified clothing, where the clothing could be Hoodie, Shirt, or T-shirt. Image AID-4 shows three sets of outfit, including multiple possible types of clothing: Chiffon, Jacket, Shirt, Suit, and T-shirt.

**Fig. 1** A depiction of NC instances versus clean instances in the Clothing1M dataset. This figure is divided into three colored sections: orange, yellow and green. The images in the orange section depict OOD instances. The images in the yellow section depict ambiguous ID (AID) instances. The images in the green section depict clean instances, i.e. images with correct labels. The 14 given labels in Clothing1M are: T-Shirt, Shirt, Knitwear, Chiffon, Sweater, Hoodie, Windbreaker, Jacket, Downcoat, Suit, Shawl, Dress, Vest, Underwear

role of image content in these NC instances. Ambiguous ID instances, especially those that do not fit neatly into a single label class (e.g. due to vague or incomplete object presentation, occlusion, etc.), could still hurt performance even if they are correctly labeled. Although incorrectly labeled by definition, OOD instances could still have similar visual features that are present in images of certain label classes, which may distort the learned feature space (and so hurt performance), especially if they could be confused even by humans for some of the given label classes.

More recent methods avoid this over-simplification by incorporating various data sampling techniques based on image content (Albert et al., 2022; Guo et al., 2018; Han et al., 2019; Lee et al., 2018; Tu et al., 2020; Xu et al., 2021; Yao et al., 2021), many of which use a common underlying idea that NC instances can be identified via entropy maximization (Albert et al., 2022; Chan et al., 2021; Kirsch et al., 2021; Macêdo & Ludermir, 2021a; Macêdo et al., 2021b,c; Yao et al., 2021; Yu & Aizawa, 2019). Informally, the most easily identifiable NC instances are precisely those with near-maximum entropy. To make sense of this information-theoretic notion of (Shannon) entropy, we implicitly assume[1] that, with respect to a trained classifier, instances have prediction vectors that are valid as parameter vectors of multinomial distributions. Moreover, we interpret the *prediction* of an instance probabilistically to be a random variable with this multinomial distribution. The *entropy* of an instance is then simply the entropy of its prediction. Since the discrete uniform distribution has maximum entropy by definition, it follows that the most easily identifiable NC instances are precisely those instances whose predictions are "almost" uniformly distributed. Thus, entropy maximization is the idea of categorizing an instance as an NC instance if its entropy exceeds a certain threshold.

However, there is a fundamental limitation of using entropy maximization: *Not all NC instances have predictions that are "almost" uniformly distributed.* Consider the hypothetical example of a web image dataset with 10 classes, consisting of 5 animal classes and 5 plant classes. Let $x$ be an instance whose prediction score for each animal class is

---

[1] This is a mild assumption, since softmax output layers are frequently used.

**Table 1** The $(\alpha, \beta)$-generalized KL divergences, KL divergences, normalized entropies, and prediction vectors, for the images shown in Fig. 1. (We used $p = \left[\frac{1}{k}, \ldots, \frac{1}{k}\right] \in \mathbb{R}^k$, $\alpha = 0.7$ and $\beta = 0.03$ for our $(\alpha, \beta)$-generalized KL divergence, where $k = 14$ is the number of classes in Clothing1M.)

| Image | $(\alpha, \beta)$-generalized KL divergence | KL divergence | normalized entropy | prediction vector |
|---|---|---|---|---|
| OOD-1 | 0.267 | 0.215 | 0.946 | [0.08, 0.11, 0.05, 0.05, 0.05, 0.12, 0.02, 0.06, 0.03, 0.04, 0.08, 0.13, 0.15, 0.04] |
| AID-1 | 0.265 | 0.240 | 0.940 | [0.09, 0.02, 0.07, 0.04, 0.04, 0.05, 0.05, 0.04, 0.02, 0.14, 0.14, 0.14, 0.08, 0.07] |
| C-1 | **-0.422** | 0.268 | 0.937 | [0.15, 0.11, 0.05, 0.10, 0.04, 0.03, 0.13, 0.06, 0.06, 0.08, 0.03, 0.11, 0.02, 0.03] |
| OOD-2 | 0.296 | 0.285 | 0.917 | [0.05, 0.06, 0.05, 0.03, 0.07, 0.06, 0.05, 0.07, 0.02, 0.03, 0.25, 0.11, 0.07, 0.09] |
| AID-2 | 0.195 | 0.313 | 0.941 | [0.02, 0.10, 0.07, 0.01, 0.06, 0.07, 0.11, 0.08, 0.05, 0.05, 0.09, 0.09, 0.04, 0.03, 0.16] |
| OOD-3 | 0.237 | 0.319 | 0.927 | [0.11, 0.09, 0.04, 0.02, 0.06, 0.05, 0.04, 0.05, 0.02, 0.02, 0.10, 0.19, 0.10, 0.09] |
| AID-3 | 0.221 | 0.482 | 0.910 | [0.08, 0.07, 0.10, 0.00, 0.11, 0.02, 0.06, 0.08, 0.02, 0.05, 0.22, 0.09, 0.05, 0.06] |
| OOD-4 | 0.332 | 0.509 | 0.872 | [0.12, 0.05, 0.11, 0.01, 0.06, 0.05, 0.05, 0.05, 0.04, 0.06, 0.03, 0.02, 0.07, 0.29] |
| AID-4 | 0.218 | 0.725 | 0.884 | [0.06, 0.06, 0.07, 0.05, 0.05, 0.16, 0.05, 0.07, 0.01, 0.09, 0.00, 0.10, 0.21, 0.01] |
| C-2 | **-2.665** | 21.275 | 0.000 | [0.00, 0.00, 0.00, 0.00, 0.00, 0.00, 0.00, 0.00, 0.00, 0.00, 0.00, 0.00, 0.00, 1.00] |

The table rows are arranged according to the KL divergences in ascending order. Normalized entropy refers to entropy divided by the maximum possible entropy ($\log k$). For prediction vectors, the non-dominant entries with values lesser than $\frac{1}{k} - \beta$ are highlighted in red. Note that there is no possible threshold for both KL divergence and normalized entropy that would distinguish clean instances from NC instances across all ten images in Fig. 1, since image C-1 has low KL divergence and high normalized entropy, while image C-2 has high KL divergence and low normalized entropy. In contrast, it is possible to distinguish non-NC instances from NC instances by checking whether their $(\alpha, \beta)$-generalized KL divergence is negative (see boldfaced values) or non-negative, respectively

0.20, and whose prediction scores for all plant classes are 0.00. Clearly, $x$ could be interpreted as an NC instance that is predicted to be related to the 5 animal classes with equal uncertainty, but predicted to be unrelated to any of the plant classes. Next, consider another instance $x'$, whose prediction score for the first animal class is 0.55, and whose prediction scores for all remaining 9 classes are 0.05 each. In contrast, $x'$ is naturally not an NC instance, since it fits well into exactly one class. However, the normalized[2] entropy of $x$ is $\frac{\log 5}{\log 10} \approx 0.699$, while the normalized entropy of $x'$ is $\approx 0.728$. If the threshold is less than 0.728, then $x'$ would be misclassified as an NC instance. If the threshold is greater than 0.699, then $x$ would be misclassified as a non-NC instance. Hence, this implies that *all* possible thresholds when using normalized entropy to identify NC instances would miscategorize at least one of $x, x'$. For more concrete examples, see Table 1 for NC instances and non-NC instances (in Clothing1M) that cannot be distinguished using thresh-olds on their normalized entropies. As these examples reveal, entropy maximization is inherently inadequate for identifying NC instances.

*Beyond entropy maximization.* By definition, entropy maximization is equivalent to the minimization of the Kullback–Leibler (KL) divergence. Building upon this observation, we introduce $(\alpha, \beta)$-generalized KL divergence, a new generalization of KL divergence that is well-suited for identifying NC instances. In essence, we are extending this KL divergence minimization idea to the case of minimizing the "divergence" from $p$ to $q$, relative to those dominant entries in the parameter vector of $q$. Here, the precise meaning of "dominant" depends on our hyperparameters $\alpha$ and $\beta$, whose values we can adjust. (Intuitively, an entry is dominant if it is not small.) By using $(\alpha, \beta)$-generalized KL divergence, we are able to not only identify NC instances whose prediction vectors have entries that are all dominant (i.e. instances with "almost" uniformly distributed predictions), but also identify additional NC instances whose prediction vectors have less dominant entries. Thus, $(\alpha, \beta)$-generalized

---

[2] Normalized entropy refers to entropy divided by the maximum possible entropy ($\log k$). Here, $\log 5$ is the entropy of $x$, and $\log 10$ is its maximum possible entropy.

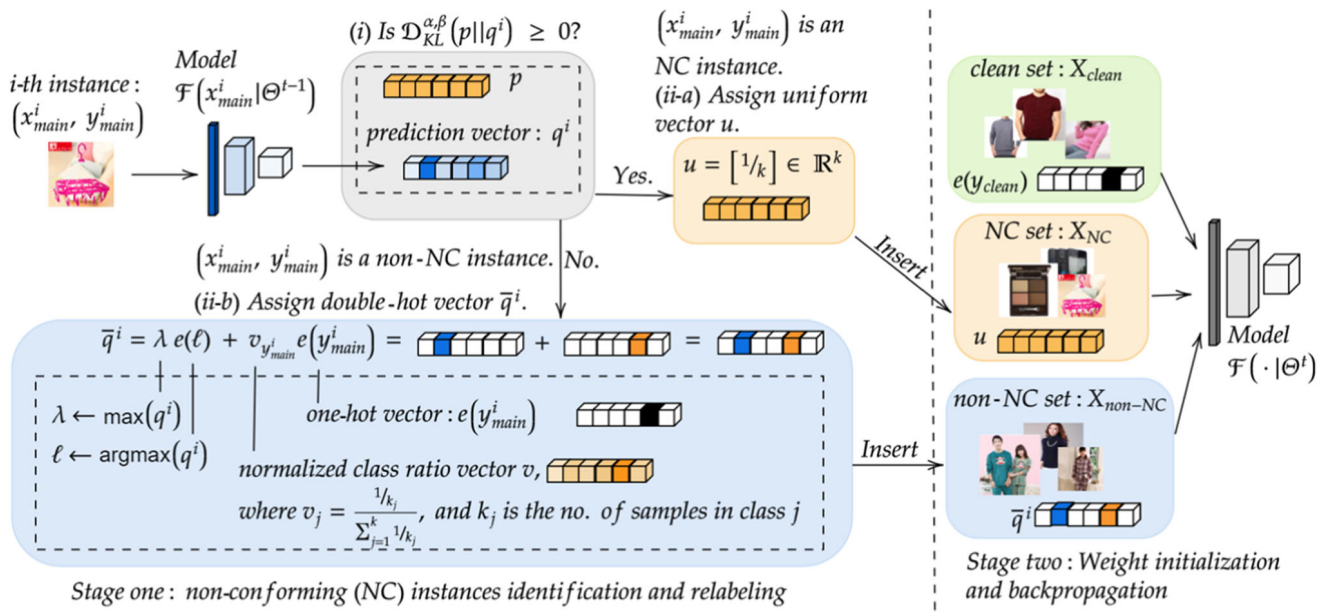

**Fig. 2** An overview of our GENKL framework in iteration $t$. GENKL has two stages; we iterate between the two stages until the model converges. Stage one includes two critical components: (i) identification of NC instances using $(\alpha, \beta)$-generalized KL divergence; and (ii) relabeling of NC and non-NC instances using soft labels. In component (i), given the $i$-th instance $(x^i_{main}, y^i_{main})$, we obtain its prediction vector $q^i$ from the model $\mathcal{F}(x^i_{main} | \Theta^{t-1})$. Next, we compute the $(\alpha, \beta)$-generalized KL divergence $\mathcal{D}^{\alpha,\beta}_{KL}(p\|q^i)$, where $p$ is a uniform-like vector (see Sect. 3.4

for its definition), then identify our $i$-th instance as an NC instance if $\mathcal{D}^{\alpha,\beta}_{KL}(p\|q^i) \geq 0$, and as a non-NC instance otherwise. In component (ii), there are two scenarios: (ii-a) If $(x^i_{main}, y^i_{main})$ is an NC instance, then assign to it a uniform vector $[\frac{1}{k}, \ldots, \frac{1}{k}]$ as its soft label; and (ii-b) If $(x^i_{main}, y^i_{main})$ is a non-NC instance, then assign to it a double-hot vector $\bar{q}^i$ as its soft label. In stage two, an initialized model is trained on $X_{non-NC}$, $X_{NC}$ with their respective soft labels, and $X_{clean}$ with its given labels. Full details can be found in Sect. 3.5

KL divergence directly addresses the fundamental limitation of using entropy maximization.

*Robust training with generalized KL divergence.* To improve the robustness of training classifiers on web image datasets with both NC instances and non-NC instances with label noise, we propose GENKL, a general training framework based on our $(\alpha, \beta)$-generalized KL divergence. There are two stages in GENKL. For stage one, there are two key components: (i) identification of NC instances using $(\alpha, \beta)$-generalized KL divergence; and (ii) relabeling of NC and non-NC instances using soft labels. For stage two, we perform weight initialization, then carry out the usual training on the relabeled instances and clean instances. By iteratively alternating between the two stages, GENKL is robust to both NC instances, and label noise in the input data. See Fig. 2 for an overview of the GENKL framework. Our experiments on web image datasets show that GENKL is able to achieve state-of-the-art (SOTA) accuracies.

Our main contributions are given as follows:

- We propose a new generalized KL divergence, $\mathcal{D}^{\alpha,\beta}_{KL}(p\|q)$, which is well-suited for identifying NC instances.
- We prove theoretical properties of $\mathcal{D}^{\alpha,\beta}_{KL}(p\|q)$, whose proofs are provided in the Appendix.

- We propose GENKL, a training framework that is robust to NC instances in training data.
- In our experiments on web image datasets, Clothing1M (Xiao et al., 2015), Food101/ Food101N (Bossard et al., 2014; Lee et al., 2018) and mini WebVision 1.0 (Li et al., 2017), we achieved new SOTA accuracies 81.34%, 85.73%, and 78.99%/92.54% (top-1/top-5), respectively.

## 2 Related Work

### 2.1 Dealing with NC Instances

Existing approaches for dealing with NC instances can be broadly categorized into two types: (i) treating NC instances more generally as instances with label noise; and (ii) identifying and alleviating the effects of NC instances. Intuitively, although instances with label noise include unambiguous ID instances with incorrect labels, which are not NC instances, any method that tackles the problem of label noise would naturally tackle the sub-problem of OOD instances, which are NC instances with incorrect labels.[3] See Fig. 3 for a Venn diagram that illustrates the relationship between NC instances and instances with label noise. (See also Sect. 3.1 for a rigorous treatment of NC instances, including how they relate to label assignment.)

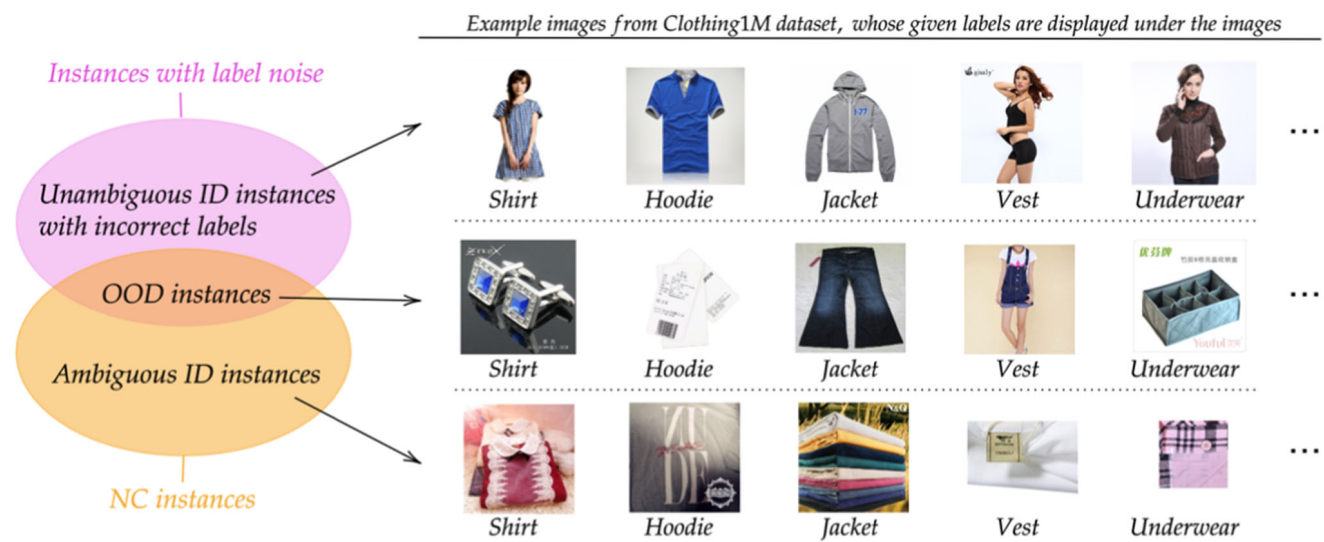

**Fig. 3** A Venn diagram that illustrates the relationship between NC instances and instances with label noise. For each sub-category in this Venn diagram, some example images are shown from the Clothing1M dataset

For the first type, existing works can be categorized based on their use of (i) estimators for noise transition matrices, (ii) robust loss functions, (iii) regularization techniques, and (iv) other specialized model architectures. An accurate estimate of the noise transition matrix is vital to many applications, because the matrix can not only model the label corruption process for the dataset, but also can be used to infer the clean class posterior probabilities of the dataset (Han et al., 2020). There are numerous methods that estimate the noise transition matrix, which is then used either to alleviate the effect of label noise (Sukhbaatar & Fergus, 2014; Xiao et al., 2015; Han et al., 2018), or to perform label correction (Hendrycks et al., 2018; Patrini et al., 2017; Xia et al., 2020, 2019). Various loss functions have been proposed (Ghosh et al., 2017; Lyu & Tsang, 2019; Ma et al., 2020; Song et al., 2019; Wang et al., 2019; Hendrycks et al., 2018; Patrini et al., 2017; Xia et al., 2019; Yao et al., 2020), which are specifically designed to be robust to noisy data. Also, regularization techniques (Ioffe & Szegedy, 2015; Jenni & Favaro, 2018; Krogh & Hertz, 1991; Pereyra et al., 2017; Shorten & Khoshgoftaar, 2019; Srivastava et al., 2014; Zhang et al., 2017) are used to increase the generalization capability of a trained model. Such techniques are frequently combined with other methods. Finally, some methods do not fit neatly into the first three sub-categories, and instead use newly proposed specialized model architectures (Wang et al., 2021; Peng et al., 2020; Yang et al., 2021; Tu et al., 2020; Yao et al., 2018). CAN (Yao et al., 2018) is a contrastive-additive noise network that has a contrastive layer to estimate a quality embedding space, and an additive layer for estimation aggregation. AFM (Peng et al., 2020) introduces a training block that suppresses mislabeled data via grouping and self-attention; see also (Xu et al., 2022). SOMNet (Tu et al., 2020) groups images and their proposed regions-of-interest (ROIs) from the same category into bags, and thereafter, weights are assigned to the bags based on their discriminative scores with the nearest clusters. These methods of the first type are designed to deal with label noise, and do not specifically target the challenge of NC instances. Hence, these methods have limited impact on web image datasets that have a non-negligible number of NC instances.

For the second type, numerous works identify NC instances using entropy-based methods. Many of them use entropy maximization to identify OOD instances (Chan et al., 2021; Kirsch et al., 2021; Macêdo & Ludermir, 2021a; Macêdo et al., 2021b,c). Other works use divergences closely related to KL divergence to identify NC instances. For example, Jo-SRC (Yao et al., 2021) uses Jensen–Shannon divergence to separate clean instances from OOD and noisy ID instances. It then compares the consistency of the predictions of each non-clean instance from multiple views (via data augmentation) to determine whether the instance is OOD or noisy ID. Another example is DSOS (Albert et al., 2022), which first separates non-clean instances from clean instances via collision entropy, then uses beta mixture models to identify noisy ID and OOD instances, and applies label correction to improve classification accuracy. It should be noted that in Albert et al. (2022), the authors hypothesized that DSOS might not perform as well if most of the non-clean instances are OOD rather than noisy ID, which may limit the effectiveness of DSOS on some web image datasets, e.g. Clothing1M.

---

[3] In particular, methods that deal with label noise typically have the implicit assumption that the given label for any instance must, with full certainty, be either correct or incorrect. Such an assumption omits the possibility that an instance could be ambiguous, whereby it is not clear what a *single* "correct" label would be, especially when there are multiple possible labels that could potentially be deemed "correct". (See Sect. 3.1 for a rigorous definition of what "ambiguous instance" means precisely.) Consequently, under such an assumption, where ambiguous ID instances are *assumed* to not exist, the remaining NC instances (i.e. OOD instances), together with the unambiguous ID instances with incorrect labels, are then treated by these methods in a general manner as instances with label noise.

## 2.2 Generalizations of KL Divergence and other divergences

**KL divergence** (also called **relative entropy**) is a non-negative, unbounded divergence between two stochastic vectors $p$ and $q$, defined as follows:

$$\mathcal{D}_{\mathrm{KL}}(p\|q) = \sum_{j=1}^{k} p_j \log(p_j) - \sum_{j=1}^{k} p_j \log(q_j)$$

Note that KL divergence is asymmetric, i.e. in general, $\mathcal{D}_{\mathrm{KL}}(p\|q) \neq \mathcal{D}_{\mathrm{KL}}(q\|p)$. There are multiple variants of KL divergence and other divergences:

- **Jeffreys divergence** (Jeffreys, 1998) is defined by

  $$\mathcal{D}_{\mathrm{Jeff}}(p\|q) := \frac{1}{2}\mathcal{D}_{\mathrm{KL}}(p\|q) + \frac{1}{2}\mathcal{D}_{\mathrm{KL}}(q\|p),$$

  which is a symmetric analog of KL divergence.
- **Jensen–Shannon (JS) divergence** is defined by

  $$\mathcal{D}_{\mathrm{JS}}(p\|q) := \frac{1}{2}\mathcal{D}_{\mathrm{KL}}(p\|\tfrac{p+q}{2}) + \frac{1}{2}\mathcal{D}_{\mathrm{KL}}(q\|\tfrac{p+q}{2}),$$

  which is another symmetric analog of KL divergence. This divergence is bounded within [0, 1] if logarithms are taken over base 2 (Lin, 1991).
- **Decision Cognizant (DC) KL divergence** was first introduced in Ponti et al. (2017). Suppose $\arg\max(p) = s$, $\arg\max(q) = t$, and define the set $\Lambda := \{1, \ldots, k\} \backslash \{s, t\}$. Then this divergence is defined by

  $$\mathcal{D}_{\mathrm{DC}}(p\|q) := \sum_{j \in \{s,t\}} q_j \log \frac{q_j}{p_j} + \Big(\sum_{j \in \Lambda} q_j\Big) \log \frac{\sum_{j \in \Lambda} q_j}{\sum_{j \in \Lambda} p_j}.$$

  Similar to KL divergence, this DC KL divergence is non-negative, unbounded and asymmetric. The main difference for this divergence is that the contributions from minority classes are reduced, which is what (Ponti et al., 2017) refers to as being "decision cognizant".

- **Delta divergence** (Kittler & Zor, 2018): Let $\arg\max(p) = s$, let $\arg\max(q) = t$, and define the index set $\Lambda := \{1, \ldots, t\} \backslash \{s, t\}$. Then Delta divergence is defined by

  $$\mathcal{D}_{\Delta}(p\|q) := \frac{1}{2}\Big[\sum_{j \in \{s,t\}} |p_j - q_j| + |p_\Lambda - q_\Lambda|\Big],$$

  where $p_\Lambda = \sum_{j \in \Lambda} p_j$ and $q_\Lambda = \sum_{j \in \Lambda} q_j$. This divergence is non-negative, bounded, symmetric and effectively groups non-dominant entries into a single class.

**Rényi entropy** generalizes the usual notion of (Shannon) entropy, and it is defined by

$$H_\alpha(p) := \frac{1}{1-\alpha} \log\Big(\sum_{j=1}^{k} p_i^\alpha\Big),$$

where $\alpha \geq 0, \alpha \neq 1$. Note that Rényi entropy becomes Shannon entropy in the limit as $\alpha \to 1$.

**Collision entropy** (Albert et al., 2022) is a popular special case ($\alpha = 2$) of Rényi entropy, given by

$$H_2(p) := -\log\Big(\sum_{j=1}^{k} p_j^2\Big).$$

## 3 Proposed Method

This section introduces GENKL, an iterative training framework robust to both NC instances and instances with label noise. We first begin in Sect. 3.1 with a rigorous definition of NC instances. Next, we cover the preliminaries in Sect. 3.2, followed by a formal introduction to our $(\alpha, \beta)$-generalized KL divergence $\mathcal{D}_{\mathrm{KL}}^{\alpha,\beta}(p\|q)$ in Sect. 3.3. Section 3.4 describes the usage of $(\alpha, \beta)$-generalized KL divergence. We then build upon this $(\alpha, \beta)$-generalized KL divergence, and give full algorithmic details of our proposed GENKL framework in Sect. 3.5. Finally, in Sect. 3.6, we give further details on "double-hot vectors", an important ingredient of our GENKL framework.

### 3.1 What Exactly are NC Instances?

To rigorously define what NC instances are, it is helpful to think of the dataset annotation process. In particular, the annotation process of "assigning a single class label to an input image" can be interpreted in terms of solving an object detection task.

Imagine a two-step object detection process:

- Step 1: We locate the objects in the input image, which correspond to regions of the image (specified by bounding boxes) whose content is distinguishable from the background; this is commonly known as generic object detection (Liu et al., 2020); cf. Maaz et al. (2022). For convenience, let $\mathcal{O}$ denote the set of located objects.
- Step 2: Given a fixed set $\mathcal{L}$ of object class labels, we shall try to assign each object in $\mathcal{O}$ to a label in $\mathcal{L}$. We shall assume that an object is assigned a label $y \in \mathcal{L}$, only if all salient features of the object class associated to $y$ are present/detected in the bounding box of the object. It is possible that some objects in $\mathcal{O}$ cannot be assigned any label in $\mathcal{L}$, in which case, we consider such objects to be *out-of-distribution* (OOD). Objects that are assigned labels in $\mathcal{L}$ are called *in-distribution* (ID). If an ID object in $\mathcal{O}$ could be assigned multiple labels in $\mathcal{L}$, then we say it is an *ambiguous* ID object.

Based on the located objects in $\mathcal{O}$, together with the labels assigned to these objects wherever possible, our goal is to assign a single label from $\mathcal{L}$ to represent the entire image. Although our goal is seemingly simple, there are subtle issues we have to address. Crucially, is there an obvious main object of interest? Are there multiple main objects of interest? How do we characterize objects in $\mathcal{O}$ to be main objects of interest?

Among the bounding boxes of all objects in $\mathcal{O}$, let $A$ denote the maximum possible area among these bounding boxes. Given a threshold $\eta$ (e.g., $\eta = 0.5$), we shall define an object $obj \in \mathcal{O}$ to be a *main object of interest*, if the bounding box for $obj$ has an area at least $\eta A$. Let $\mathcal{O}_{\text{interest}}$ denote the subset of $\mathcal{O}$ consisting of main objects of interest.[4]

For a single label $y \in \mathcal{L}$ to accurately represent an input image, we require every object in $\mathcal{O}_{\text{interest}}$ to be assigned a single label $y$. This brings us to our formal definition of NC instances:

**Definition 1** Fix a set $\mathcal{L}$ of object class labels. For a given instance $(x, y)$ (where $x$ represents an image, and $y$ is the corresponding given label in $\mathcal{L}$, which may possibly be incorrect), define $\mathcal{O}_{\text{interest}}$ for image $x$ as above. Then we call $(x, y)$ an *NC instance*, if there is no possible single label $y' \in \mathcal{L}$ such that every object in $\mathcal{O}_{\text{interest}}$ is assigned the same label $y'$.[5]

---

[4] A more sophisticated definition for "main object of interest" is certainly possible. However, when defining NC instances, we only need the notion of "main object of interest" to be well-defined. Whichever semantics we choose to impose on this notion, our definition of NC instances would still make sense. Hence, we shall use our simplified definition of "main object of interest".

[5] Note that $y'$ can be different from $y$. In particular, an instance with an incorrect given label may not be an NC instance; see Fig. 3 for a visualization.

*Examples.* In Fig. 1, images AID-1/2/3/4 and OOD-1/2/3/4 depict NC instances. The main object(s) of interest in images AID-1/2/3/4 are ambiguous ID objects, and the main object(s) of interest in images OOD-1/2/3/4 are OOD objects.

## 3.2 Preliminaries

Throughout, given any vector $v$, we shall let its $j$-th entry be denoted by $v_j$. (We shall always reserve subscripts on vectors to refer to their entries.) For any dataset $\mathcal{D} = \{(x^i, y^i)\}_{i=1}^N$ with $N$ instances, we use the convention that the $i$-th instance is the pair $(x^i, y^i)$, where $x^i$ is the $i$-th feature vector/image. The corresponding label $y^i$ is an integer $j \in \{1, \ldots, k\}$. Let $e(y^i) = e(j)$ be the one-hot vector whose $j$-th entry is 1, and remaining entries are 0. For convenience, let $X_{\mathcal{D}} := (x^i)_{i=1}^N$ (resp. $Y_{\mathcal{D}} := (y^i)_{i=1}^N$) be the sequence of feature vectors/images (resp. sequence of labels) for $\mathcal{D}$. A neural network model in iteration $t$ is denoted by $\mathcal{F}(\cdot | \Theta^t)$, where $\Theta^t$ represents the model weights in iteration $t$. Assume that for every input $x^i$, the output $q^i = \mathcal{F}(x^i | \Theta^t)$ is a stochastic vector. For convenience, we define the *entropy* of a stochastic vector $p$ to be $H(p) := -\sum_{j=1}^k p_j \log(p_j)$. Analogously, for stochastic vectors $p$ and $q$, we define the *cross-entropy* from $p$ to $q$ to be $H(p, q) := -\sum_{j=1}^k p_j \log(q_j)$. In both $H(p)$ and $H(p, q)$, we use the usual convention that $0 \log 0 = 0$. Later, we shall abuse notation and use $H(p, q)$ for the case when $p$ is a non-stochastic vector. In this case, $H(p, q)$ is defined using the same formula.

## 3.3 $(\alpha, \beta)$-generalized KL Divergence

Given any two stochastic vectors $p$ and $q$ of length $k$, and given real values $\alpha, \beta$ satisfying $\alpha > 0$ and $0 \leq \beta \leq \frac{1}{k}$, we shall define the $(\alpha, \beta)$-generalized KL divergence from $p$ to $q$ as follows:

$$\mathcal{D}_{\text{KL}}^{\alpha, \beta}(p \| q) = \alpha \sum_{j=1}^k p_j \log(p_j) - \sum_{j=1}^k p_j \cdot \mathbb{1}_{q_j \geq \frac{1}{k} - \beta} \log(q_j). \tag{1}$$

Here, $\mathbb{1}_{q_j \geq \frac{1}{k} - \beta}$ is the indicator function that returns 1 if $q_j \geq \frac{1}{k} - \beta$, and returns 0 otherwise. Succinctly, we can write (1) as:

$$\mathcal{D}_{\text{KL}}^{\alpha, \beta}(p \| q) = -\alpha H(p) + H(^\beta p, q), \tag{2}$$

where $^\beta p := [p_1 \mathbb{1}_{q_1 \geq \frac{1}{k} - \beta}, \ldots, p_k \mathbb{1}_{q_k \geq \frac{1}{k} - \beta}]$. Note that we can easily compute $^\beta p$ by replacing those entries in $p$ that are less than $\frac{1}{k} - \beta$ with the value 0. In both (1) and (2), we use the usual convention that $0 \log 0 := 0$.

There are two hyperparameters in $\mathcal{D}_{\mathrm{KL}}^{\alpha,\beta}(p\|q)$: $\alpha$ and $\beta$. Note that when $\alpha = 1$ and $\beta = \frac{1}{k}$, our $(\alpha,\beta)$-generalized KL divergence coincides exactly with the usual KL divergence. Informally, the first term $-\alpha H(p)$ is negative, where $\alpha > 0$ is a hyperparameter that controls "how negative" this weighted "entropy" term is. The second hyperparameter $\beta$ appears in the second term $H(^{\beta}p, q)$, which is a positive "cross-entropy" term. Intuitively, $\beta$ controls the threshold of what it means to be a non-dominant entry. Given a stochastic vector $q$ of length $k$, we say that its $i$-th entry $q_i$ is *dominant* if $q_i \geq \frac{1}{k} - \beta$, and *non-dominant* otherwise. Thus, we are effectively computing the cross-entropy only for the dominant entries of $q$, where the contributions of the non-dominant entries of $q$ are ignored.

**Theorem 2** *Let $p$ and $q$ be two stochastic vectors of length $k \geq 2$. Let $\alpha > 0$ and $\beta \in [0, \frac{1}{k}]$. Then $(\alpha,\beta)$-generalized KL divergence $\mathcal{D}_{\mathrm{KL}}^{\alpha,\beta}(p\|q)$ is bounded as follows:*

*(i)*

$$
\mathcal{D}_{\mathrm{KL}}^{\alpha,\beta}(p\|q) \geq \begin{cases} 0, & \text{if } \beta = \frac{1}{k}, \, 0 < \alpha \leq 1; \\ (1-\alpha)\log k, & \text{if } \beta = \frac{1}{k}, \, \alpha > 1; \quad (3) \\ -\alpha \log k, & \text{if } 0 \leq \beta < \frac{1}{k}. \end{cases}
$$

- *If $\beta = \frac{1}{k}$ and $0 < \alpha \leq 1$, then equality holds if and only if $q = p$ and $p$ is a one-hot vector.*
- *If $\beta = \frac{1}{k}$ and $\alpha > 1$, then equality holds if and only if $p = q$ and $p$ is a uniform vector.*
- *If $0 \leq \beta < \frac{1}{k}$, then equality holds if and only if $q$ is a one-hot vector and $p$ is a uniform vector.*

*(ii) If $\beta = \frac{1}{k}$, then $\mathcal{D}_{\mathrm{KL}}^{\alpha,\beta}(p\|q)$ is unbounded. If instead $0 \leq \beta < \frac{1}{k}$, then $\mathcal{D}_{\mathrm{KL}}^{\alpha,\beta}(p\|q) \leq \log \frac{1}{\frac{1}{k} - \beta}$.*

*A detailed proof is provided in Appendix A.*

Informally, Theorem 2 gives the full range of values that $\mathcal{D}_{\mathrm{KL}}^{\alpha,\beta}(p\|q)$ can attain, over all possible pairs of values for the hyperparameters $\alpha$ and $\beta$. In the special case when $\alpha = 1$ and $\beta = \frac{1}{k}$, $\mathcal{D}_{\mathrm{KL}}^{\alpha,\beta}(p\|q)$ is the usual KL divergence from $p$ to $q$, and Theorem 2 becomes the well-known "information" from Information Theory; see (Thomas & Joy 2006, Thm. 2.6.3).

**Theorem 3** *For any $0/1$-vector $u = [u_1, \ldots, u_k]$, $\alpha \geq 1$, and $\beta \in [0, \frac{1}{k}]$, we define*

$$
\mathbb{R}_u^{\beta} := \Big\{ (p, q) : p \text{ and } q \text{ are stochastic vectors,}
$$
$$
\mathbb{1}_{q_i \geq \frac{1}{k} - \beta} = u_i \text{ for all } 1 \leq i \leq k \Big\}.
$$

*Then for all $\alpha \geq 1$, $\beta \in [0, \frac{1}{k}]$, we have that $(\alpha, \beta)$-generalized KL divergence is piecewise convex: This means that for all $0/1$-vectors $u$, all pairs $(p, q)$, $(p', q') \in \mathbb{R}_u^{\beta}$, and all $\lambda \in [0, 1]$,*

$$
\mathcal{D}_{\mathrm{KL}}^{\alpha,\beta}(\lambda p + (1-\lambda)p' \| \lambda q + (1-\lambda)q')
$$
$$
\leq \lambda \mathcal{D}_{\mathrm{KL}}^{\alpha,\beta}(p\|q) + (1-\lambda)\mathcal{D}_{\mathrm{KL}}^{\alpha,\beta}(p'\|q').
$$

*A detailed proof is provided in Appendix B.*

It is well-known that KL divergence $\mathcal{D}_{\mathrm{KL}}(p\|q)$ is convex in the pair $(p, q)$; see (Thomas & Joy 2006, Thm. 2.7.2). Theorem 3 extends this result to the general case of $(\alpha, \beta)$-generalized KL divergence, where instead of convexity, we have piecewise convexity for $\mathcal{D}_{\mathrm{KL}}^{\alpha,\beta}(p\|q)$.

### 3.4 Identification of NC Instances

Later, when we introduce GenKL in Sect. 3.5, we shall be considering two disjoint datasets $\mathcal{D}_{\mathrm{main}}$ and $\mathcal{D}_{\mathrm{clean}}$ with the same set of label classes, where $\mathcal{D}_{\mathrm{main}}$ has both NC instances and instances with label noise, while $\mathcal{D}_{\mathrm{clean}}$ is assumed to contain only non-NC instances that are clean (i.e., all labels are correct). We do *not* assume that the feature vectors/images of $\mathcal{D}_{\mathrm{main}}$ and $\mathcal{D}_{\mathrm{clean}}$ are sampled from the same distribution. We shall use labels "main" and "clean" to indicate membership in the respective datasets; for example, $X_{\mathrm{main}}$ refers to the sequence $X_{\mathcal{D}_{\mathrm{main}}}$, while $(x_{\mathrm{clean}}^i, y_{\mathrm{clean}}^i)$ refers to the $i$-th instance of $\mathcal{D}_{\mathrm{clean}}$.

When using $\mathcal{D}_{\mathrm{KL}}^{\alpha,\beta}(p\|q)$ to identify NC instances, the vector $q$ we used is a prediction vector from a trained neural network. To capture the variance of the predictions, we use multiple "uniform-like" vectors for $p$, which is defined as follows: First, we sample a vector $\hat{p}$, where each entry $\hat{p}_j$ is sampled from the normal distribution $\mathcal{N}(\frac{1}{k}, \sigma^2)$. (When $\sigma = 0$, $\hat{p}$ is a uniform vector.) If the entries of $\hat{p}$ are all non-negative, then we normalize the vector $\hat{p}$ to generate the "uniform-like" vector $p$, where $p_j = \frac{\hat{p}_j}{\sum_{i=1}^{k} \hat{p}_j}$. Let $P$ be the set of uniform-like vectors $p$ obtained via this sampling process. Given an instance with prediction vector $q$, we say that this instance is an *NC instance* if $\mathcal{D}_{\mathrm{KL}}^{\alpha,\beta}(p\|q) \geq 0$ for any $p$ in $P$.

### 3.5 GenKL Framework

Our framework GenKL comprises two stages. In stage one, we identify NC and non-NC instances in $\mathcal{D}_{\mathrm{main}}$, and generate respective soft labels for both types of instances. In stage two, we perform training on $\mathcal{D}_{\mathrm{main}}$ with its newly assigned labels and $\mathcal{D}_{\mathrm{clean}}$ with its given labels (usual training). We iterate between stage one and stage two, until the model converges.

*Stage one.* There are two critical components in stage one: (i) identification of NC instances using $(\alpha, \beta)$-generalized

KL divergence; and (ii) relabeling of NC and non-NC instances using soft labels.

For component (i), we obtain the set $Q$ of prediction vectors for all instances in $\mathcal{D}_{\text{main}}$ from our model $\mathcal{F}(\cdot \mid \Theta^{t-1})$. Using our $(\alpha, \beta)$-generalized KL divergence, we then partition $X_{\text{main}}$ into two sub-sequences: $X_{\text{NC}}$ and $X_{\text{non-NC}}$, comprising the images of NC instances and non-NC instances, respectively. In particular, our $(\alpha, \beta)$-generalized KL divergence allows the identification of NC instances whose predictions are not "almost" uniformly distributed.

For component (ii), the uniform vector $u := [\frac{1}{k}, \ldots, \frac{1}{k}]$ is assigned to every image in $X_{\text{NC}}$ as its soft label. This makes the model more likely to have uniform-like prediction vectors for NC instances. In contrast, we assign what we call a double-hot vector, $\bar{q}^i$, as the soft label to an image in $X_{\text{non-NC}}$. The *double-hot vector* $\bar{q}^i$ of a non-NC instance $(x_{\text{main}}^i, y_{\text{main}}^i)$ is defined as follows: First, let the vector of normalized class ratios of $\mathcal{D}_{\text{pre}}$ [6] be denoted by $v$, and let the size of class $j$ in $\mathcal{D}_{\text{pre}}$ be $k_j$. This means that the $j$-th entry of $v$ is:

$$v_j := \frac{\frac{1}{k_j}}{\sum_{j=1}^{k} \frac{1}{k_j}}.$$

(This vector $v$ of normalized class ratios was previously used in Li et al. (2019) to define a weighted cross-entropy loss for imbalanced datasets.) Let $\lambda' = v_{y_{\text{main}}^i}$, $\lambda = \max(q^i)$, and $\ell = \arg \max q^i$. In words, $\lambda'$ is the class ratio for the class label of $x_{\text{main}}^i$, while $\lambda$ is the maximum value of the entries in the prediction vector $q^i$, corresponding to an entry with index $\ell$. Then $\bar{q}^i$ is defined by:

$$\bar{q}^i := \lambda' e(y_{\text{main}}^i) + \lambda e(\ell). \tag{4}$$

Note that $\bar{q}^i$ is a weighted sum of two one-hot vectors, where the two weights $\lambda$ and $\lambda'$ do not necessarily sum to 1. Let $\bar{Q}$ denote the sequence of double-hot vectors corresponding to $X_{\text{non-NC}}$. More details on double-hot vectors, including its motivation and interpretation, can be found later in Sect. 3.6.

*Stage two.* In iteration $t$, a model is initialized and trained on $X_{\text{clean}}$ with given labels $Y_{\text{clean}}$, $X_{\text{NC}}$ with the uniform vector $u$ as the common soft label, and $X_{\text{non-NC}}$ with double-hot vectors $\bar{Q}$ as soft labels, respectively. Their respective loss functions are defined by (5), (6), and (7):

$$\mathcal{L}_{\text{clean}}((X_{\text{clean}}, Y_{\text{clean}}), \Theta^t)$$
$$= -\frac{1}{|X_{\text{clean}}|} \sum_{i=1}^{|X_{\text{clean}}|} \sum_{j=1}^{k} e(y_{\text{clean}}^i)_j \log\Big(\mathcal{F}(x_{\text{clean}}^i \mid \Theta^t)_j\Big); \tag{5}$$

[6] We define $\mathcal{D}_{\text{pre}}$ to be the dataset that the initial model $\mathcal{F}(\cdot \mid \Theta^0)$ in the first iteration is pretrained on. Note that $\mathcal{D}_{\text{pre}}$ could be $\mathcal{D}_{\text{clean}}$, or $\mathcal{D}_{\text{main}} \cup \mathcal{D}_{\text{clean}}$. See Sect. 5.3 for more implementation details.

$$\mathcal{L}_{\text{non-NC}}((X_{\text{non-NC}}, \bar{Q}), \Theta^t)$$
$$= -\frac{1}{|X_{\text{non-NC}}|} \sum_{i=1}^{|X_{\text{non-NC}}|} \log\left(\sum_{j=1}^{k} \bar{q}_j^i \mathcal{F}(x_{\text{non-NC}}^i \mid \Theta^t)_j\right); \tag{6}$$

$$\mathcal{L}_{\text{NC}}((X_{\text{NC}}, u), \Theta^t)$$
$$= -\frac{1}{|X_{\text{NC}}|} \sum_{i=1}^{|X_{\text{NC}}|} \sum_{j=1}^{k} \frac{1}{k} \log\Big(\mathcal{F}(x_{\text{NC}}^i \mid \Theta^t)_j\Big). \tag{7}$$

The overall loss function is defined by:

$$\mathcal{L}_{\text{all}}((X_{\text{clean}}, Y_{\text{clean}}), (X_{\text{non-NC}}, \bar{Q}), (X_{\text{NC}}, u), \Theta^t)$$
$$= \omega_1 \mathcal{L}_{\text{clean}}((X_{\text{clean}}, Y_{\text{clean}}), \Theta^t)$$
$$+ \omega_2 \mathcal{L}_{\text{non-NC}}((X_{\text{non-NC}}, \bar{Q}), \Theta^t)$$
$$+ \omega_3 \mathcal{L}_{\text{NC}}((X_{\text{NC}}, u), \Theta^t), \tag{8}$$

where $\omega_1$, $\omega_2$ and $\omega_3$ are hyperparameters that represent the weightage of the contributions of $X_{\text{clean}}$, $X_{\text{NC}}$ and $X_{\text{non-NC}}$ to the overall loss.

The model weights update process is given by SGD $(\mathcal{L}_{\text{all}}(X_{\text{clean}}, Y_{\text{clean}}), (X_{\text{non-NC}}, \bar{Q}), (X_{\text{NC}}, u), \Theta^t))$ in iteration $t$. A trained model is returned at the end of stage two, which would produce the prediction vectors for $\mathcal{D}_{\text{main}}$ in the next iteration. See Algo. 1 for full algorithmic details.

### 3.6 More on Double-Hot Vectors

The role of double-hot vectors in loss optimization during training can naturally be viewed from the lens of information theory. The key idea we use is that information content can be assigned to events of random processes. Recall that for any event $A$ with corresponding probability $\Pr(A) = p$, the *information content* of $A$ is defined to be $-\log p$, which intuitively quantifies how "surprising" the occurrence of that event would be.

In our paper, the prediction of an instance is defined to be a random variable, where the set of possible outcomes is $\{1, \ldots, k\}$, i.e. the set of all possible labels. For convenience, an outcome of the prediction shall be called a *predicted label*. This means if $q$ is the prediction vector of an instance, then each entry $q_j$ is the corresponding probability that the predicted label is $j$. Thus, the usual cross entropy loss of an instance can be interpreted as the information content of the event that the predicted label is the given label. For a clean instance, this cross entropy loss (see (5)) becomes the information content of the event that the predicted label is the correct label.

For a given non-NC instance, in addition to the usual prediction vector $q$, we also have a double-hot vector $\bar{q}$. Let $A$

---

**Algorithm 1** Pseudocode for GENKL.

---

**Require:** Pre-trained weights $\Theta^0$, $num\_iters$, "uniform-like" vector set $P$, $k$, $\alpha$, $\beta$, $\omega_1$, $\omega_2$, $\omega_3$.
**Ensure:** Final trained model $\mathcal{F}(\cdot \mid \Theta^{num\_iters})$

1: **for** $t$ from 1 to $num\_iters$ **do**
    ▷ Stage one: NC instances identification and relabeling.
2:   Initialize $X_{\text{NC}}$, $X_{\text{non-NC}}$, $\bar{Q}$ as empty sequences.
3:   $q^i \leftarrow \mathcal{F}(x_{\text{main}}^i \mid \Theta^{t-1})$                                                  ▷ Generate prediction vectors for $\mathcal{D}_{\text{main}}$.
4:   **for** $p \in P$ and $(x_{\text{main}}^i, y_{\text{main}}^i) \in \mathcal{D}_{\text{main}}$ **do**
5:     **for** $j \in \{1, \ldots, k\}$ **do**
6:       **if** $q_j^i < \frac{1}{k} - \beta$ **then**
7:         $p_j \leftarrow 0$                                             ▷ Set $p_j$ to be 0 if $q_j^i$ is a non-dominant entry.
8:       **end if**
9:     **end for**
      ▷ Identify NC instances.
10:    **if** $\mathcal{D}_{\text{KL}}^{\alpha, \beta}(p \| q^i) \geq 0$ **then**
11:     Insert $x_{\text{main}}^i$ into $X_{\text{NC}}$.
12:    **else**
13:     Insert $x_{\text{main}}^i$ into $X_{\text{non-NC}}$.
      ▷ Relabel non-NC instances.
14:     $\lambda \leftarrow \max(q^i), \ell \leftarrow \arg\max(q^i), \lambda' \leftarrow v_{y_{\text{main}}^i}$
15:     $\bar{q}^i \leftarrow \lambda e(\ell) + \lambda' e(y_{\text{main}}^i)$.                            ▷ Generate double-hot vector $\bar{q}^i$ for $X_{\text{non-NC}}$.
16:     Insert $\bar{q}^i$ into $\bar{Q}$.
17:    **end if**
18:   **end for**
    ▷ Stage two: weight initialization and backpropagation.
19:   Initialize weights $\Theta^t$.
20:   $\Theta^t \leftarrow \text{SGD}(\mathcal{L}_{\text{all}}((X_{\text{clean}}, Y_{\text{clean}}), (X_{\text{non-NC}}, \bar{Q}), (X_{\text{NC}}, u), \Theta^t))$     ▷ Update model weights.
21:   $\Theta^t \leftarrow \text{SGD}(\mathcal{L}_{\text{clean}}((X_{\text{clean}}, Y_{\text{clean}}), \Theta^t))$                   ▷ Fine-tune the model on $\mathcal{D}_{\text{clean}}$.
22: **end for**
23: **return** $\mathcal{F}(\cdot \mid \Theta^{num\_iters})$                                    ▷ Return final trained model.

---

be the event that the predicted label is the correct label. Naturally, we are interested in event $A$, but we also have to deal with the uncertainty that the given label may be incorrect. Informally, we can think of both vectors $q$ and $\bar{q}$ as measurements of two independent random processes that each yields information about the correct label of the given non-NC instance, and we would like to quantify the "combined" information we get from the measurements.

For an instance $(x, y)$ with given prediction vector $q$, the double-hot vector $\bar{q}$ associated to this instance has at most two non-zero entries, at indices $y$ and $\ell := \arg\max q$. (Note that $\bar{q}$ has only one non-zero entry if $y = \ell$.) Recall that the value of $\lambda'$ in (4) depends only on the class distribution of the entire dataset $\mathcal{D}_{\text{pre}}$, where $\lambda'$ is larger if the normalized class ratio for object class $y$ is larger (i.e. if class $y$ is rarer, in the case that $\mathcal{D}_{\text{pre}}$ is an imbalanced dataset). Intuitively, a given label is "more surprising" if it corresponds to a rarer class, and in this "more surprising" case, we would like to assign a larger prior belief probability that the given label is correct. Hence, $\lambda'$ can be interpreted as a measure of our prior belief that the given label is correct. The value $\lambda := q_\ell$ in (4) is by definition the probability of the most probable predicted label $\ell$. Hence, $\lambda$ can be interpreted as a measure of the model's belief that the most probable predicted label is correct. Consequently, the double-hot vector $\bar{q}$ can be interpreted as the overall relative measure of our prior belief (based on normalized class ratios) in comparison to the model's belief (based on training on the given labels), for the correctness of the given label versus the most probable predicted label.

If $\bar{q}$ is a stochastic vector, then we can define a random variable $V$ for the same set $\{1, \ldots, k\}$ of possible outcomes, such that $\Pr(V = j) = \bar{q}_j$. For convenience, an outcome of $V$ shall be called a *belief label*. Hence, each entry $\bar{q}_j$ is the corresponding probability that the belief label is $j$. Thus, we can interpret the loss value in (6) as the information content of $A \cap B$, where $A$ is the event defined as above, and $B$ is the event that the belief label is the correct label. Under the assumption that $A$ and $B$ are independent, it then follows from the law of total probability that $\Pr(A \cap B) = \sum_{j=1}^k q_j \bar{q}_j$, so the information content of event $A \cap B$ is $-\log\left(\sum_{j=1}^k q_j \bar{q}_j\right)$.

Intuitively, as we minimize the value of the loss function $\mathcal{L}_{\text{non-NC}}$ (see (6)), we are maximizing the probability that the correct label for the given non-NC instance is either the given label or the most probable prediction label, i.e. exactly one of the indices of the non-zero entries of the double-hot vector. This intuition still holds when we drop the requirement that $\bar{q}$ must be a stochastic vector, and later in our ablation study (see Table 6), we show that allowing $\bar{q}$ to be a non-stochastic vector yields better performance in our experiments.

## 4 Discussion

Not all NC instances have "almost" uniformly distributed predictions. This was the fundamental limitation of entropy maximization methods that we highlighted in Sect. 1. Through the use of $(\alpha, \beta)$-generalized KL divergence, $\mathcal{D}_{\mathrm{KL}}^{\alpha, \beta}(p\|q)$, we are able to overcome this limitation and identify more NC instances. (Later in Sect. 5.2, we report our experiment results on the NC instance identification task.)

Intuitively, the additional NC instances we identified would have prediction vectors whose entries are not all dominant, such that the values of those dominant entries are near-uniform. Since the prediction vector represents the multinomial distribution of the prediction (treated as a random variable), it then follows from our definition of NC instances (in Sect. 3.1) that the object classes corresponding to those dominant entries would have salient features that are detected in the images of these NC instances.

Consequently, for our proposed $\mathcal{D}_{\mathrm{KL}}^{\alpha, \beta}(p\|q)$ to be effective in identifying more NC instances, we require the implicit assumption that the prediction model can detect the salient features of all object classes that are present in any input image. Specifically, the $j$-th entry of the prediction vector should be a good measure of the presence of the salient features of the $j$-th object class, where the more confident the prediction model is in detecting the salient features, the larger this $j$-th entry should be.

For example, consider Image AID-2 in Fig. 1. This is an ambiguous ID image where the feature "wrinkle-resistant fabric" is present. Interestingly, this is a salient feature of several object classes: Shirt, Windbreaker, Suit, Shawl (e.g. satin shawl), and Underwear (e.g. wrinkle-resistant pyjamas), and we noticed that the corresponding prediction vector obtained for this image has high scores for these respective entries.

In our GENKL framework, after NC instances have been identified, we then relabel non-NC instances with double-hot vectors. As elaborated in Sect. 3.6, double-hot vectors represent the overall relative measure of the beliefs for the correctness of the given label versus the most probable predicted label. Note that the effectiveness of the double-hot vectors in capturing this overall relative measure would depend on the effectiveness of the prediction model in detecting salient features. Hence, our implicit assumption is not only important to the NC instance identification task, but also to the iterative training process in our GENKL framework.

To obtain a prediction model that is able to adequately detect the salient features of all object classes, there are two general approaches. The first approach is rather natural: Maximize the overall classification accuracy of the prediction model. This is based on the intuition that a well-trained model with high classification accuracy, especially on non-NC instances, would be able to detect the salient features of the object classes with high confidence. The second approach is to train a model on a dataset consisting of sufficiently many unambiguous ID instances. This would naturally be satisfied if NC instances are relatively rare, while for a dataset where NC instances are more common, it could be better to pre-train on a clean subset of the dataset. Here we are implicitly assuming that the model is able to detect salient features of object classes with higher confidence, when trained on more unambiguous ID instances.

Finally, note that our GENKL framework has multiple hyperparameters. For a comprehensive sensitivity analysis of the effects of different choices of hyperparameter values, see Sect. 5.5 in the next section.

## 5 Experiments

We first describe in Sect. 5.1 the datasets we used in our experiments. Next, we report our experiments for the identification of NC instances in Sect. 5.2, and our experiments for the classification of web images in Sect. 5.3. In Sect. 5.4, we analyze the effectiveness of each component of GENKL via an ablation study. Finally in Sect. 5.5, we provide a sensitivity analysis of the hyperparameters of our GENKL framework.

### 5.1 Datasets

*Clothing1M:* Clothing1M (Xiao et al., 2015) has over 1 million images collected from online shopping websites. There are a total of 14 clothing classes. During data curation, labels are automatically assigned based on the keywords in the text surrounding the collected images, which may be incorrect. The authors also provide an additional clean training set with 50k images, a clean validation set with 14k images, and a clean test set with 10k images.

*Food101/Food101N:* Food101 (Bossard et al., 2014) contains 101k food images collected from foodspotting.com, while Food101N (Lee et al., 2018) contains 310k images collected from Google, Bing, Yelp and TripAdvisor. Both datasets use a common taxonomy of 101 food classes. For Food101N, there are 305k images in the training set, of which 53k images have verified labels. It also has a validation set containing 5k images with verified labels.

*Mini WebVision 1.0:* The mini WebVision 1.0 (Jiang et al., 2018) dataset is a subset of the WebVision 1.0 dataset (Li et al., 2017). The training set of mini WebVision 1.0 has 66k images collected from Flickr and Google, which collectively form the first 50 classes of the larger WebVision 1.0 dataset. The test set of mini WebVision 1.0 consists of 2.5k images with verified labels.

**Table 2** Precision, recall/sensitivity, specificity, F1 score and kappa score of all methods for NC instance identification, on Clothing1M 50k clean data combined with 200 manually verified NC instances

| Methods | Precision | Recall/Sensitivity | Specificity | F1 score | Kappa score |
|---|---|---|---|---|---|
| Jo-SRC (Yao et al., 2021) | 0.181 | 0.250 | 0.976 | 0.210 | 0.191 |
| $\mathcal{D}_{\Delta}(p \| q)$ (Kittler & Zor, 2018) | 0.183 | 0.332 | 0.969 | 0.235 | 0.215 |
| DSOS (Albert et al., 2022) | 0.250 | 0.302 | 0.980 | 0.273 | 0.254 |
| $\mathcal{D}_{DC}(p \| q)$ (Ponti et al., 2017) | 0.271 | 0.282 | 0.984 | 0.274 | 0.260 |
| MSE | 0.240 | 0.382 | 0.974 | 0.292 | 0.274 |
| Normalized entropy | **0.438** | 0.306 | **0.991** | 0.355 | 0.346 |
| $\mathcal{D}_{KL}(p \| q)$ | 0.412 | 0.488 | 0.984 | 0.441 | 0.423 |
| $\mathcal{D}_{KL}^{\alpha, \beta}(p \| q)$ | 0.434 | **0.508** | 0.985 | **0.463** | **0.448** |

The reported results are averaged over a 5-fold cross validation

## 5.2 Experiments on Identification of NC Instances

This section compares the effectiveness of several divergences and methods to identify NC instances.

*Baselines.* We used 7 baselines in total, where 5 of these (Jo-SRC (Yao et al., 2021), DSOS (Albert et al., 2022), Delta divergence (Kittler & Zor, 2018), DC KL divergence (Ponti et al., 2017), and KL divergence) were introduced in Sect. 2. Our remaining two baselines are normalized entropy, which was described in the introduction, and the classic mean-squared error. Throughout, logarithms are taken over base 2, $q$ is the prediction vector of an instance $(x^i, y^i)$, and $p$ is a uniform vector, with the exception that in our method, $p$ is instead a uniform-like vector. Each method determines whether $(x^i, y^i)$ is an NC instance, as described below; see also Appendix C for more details.

- Jo-SRC (Yao et al., 2021) has two hyperparameters $\tau_{\text{clean}}$ and $\tau_{\text{OOD}}$. Let $q$ and $q'$ be two different prediction vectors of the same instance $(x^i, y^i)$ obtained under two data augmentations. If $1 - \mathcal{D}_{JS}(q \| e(y^i)) > \tau_{\text{clean}}$, and if

$$\min\{1, |\arg\max q - \arg\max q'|\} > \tau_{\text{OOD}},$$

then this instance is an NC instance.
- Delta divergence (Kittler & Zor, 2018) has a hyperparameter $\tau_{\Delta}$. If $\mathcal{D}_{\Delta}(p \| q) \leq \tau_{\Delta}$, then $(x^i, y^i)$ is an NC instance.
- DSOS (Albert et al., 2022) has two hyperparameters $\gamma$ and $\delta$. Let the output value of a beta mixture model with two components using input $q$ be $z$. If collision entropy $H_2(\frac{q + e(y^i)}{2}) \geq -\log(\gamma)$ and $z \geq \delta$, then $(x^i, y^i)$ is an NC instance.
- DC KL divergence (Ponti et al., 2017) has a hyperparameter $\tau_{DC}$. If $\mathcal{D}_{DC}(p \| q) \leq \tau_{DC}$, then $(x^i, y^i)$ is an NC instance.

- Mean Squared Error (MSE), which is defined by

$$\text{MSE}(p, q) := \frac{1}{k} \sum_{i=1}^{k} (p_i - q_i)^2,$$

has a hyperparameter $\tau_{\text{MSE}}$. If $\text{MSE}(p, q) \leq \tau_{\text{MSE}}$, then $(x^i, y^i)$ is an NC instance.
- Normalized entropy has a hyperparameter $\tau_{\text{Nor}}$. If normalized entropy is at least $\tau_{\text{Nor}}$, then $(x^i, y^i)$ is an NC instance.
- KL divergence has a hyperparameter $\tau_{KL}$. If $\mathcal{D}_{KL}(p \| q) \leq \tau_{KL}$, then $(x^i, y^i)$ is an NC instance.

*Experimental set-up.* For Clothing1M, we used the 50k clean set as our clean instances. We also manually verified 200 NC instances out of approximately 2500 instances randomly selected from the 1 million noisy dataset. We used ResNet-50 (He et al., 2016) pretrained on ImageNet for all experiments in this section. To keep the class ratios invariant, we first used stratified sampling to randomly select 10% of the 50k clean set as test data. For the remaining 90% of the instances, we used stratified 5-fold cross-validation to split the data. The validation set is randomly shuffled and then selected with the same size as the test set. We randomly split the 200 NC instances into 2 folds of equal sizes: One fold is used to augment the validation set, while the other fold is used to augment the test set. A model is trained to generate prediction vectors for the respective (augmented) validation set, which has 100 NC instances; see Appendix C for further experiment details.

We obtained the test accuracies for all methods using the hyperparameters tuned on the validation set. Recall that $P$ is our set of uniform-like vectors. For our experiments on the identification of NC instances, we used a set $P$ with two vectors, one of which is the uniform vector, and we used $\alpha = 1.0247$, $\beta = 0.0665$, and $\sigma = 0.06$.

*Evaluation metrics.* The main metrics used are F1 score and Cohen's kappa score (Feuerman & Miller, 2005). Let the number of true positives, true negatives, false positives, and false negatives be denoted by TP, TN, FP and FN, respectively. Note that the number of predicted positives (resp. predicted negatives) is given by TP+FP (resp. TN+FN).

- *Precision* is the ratio of true positives to predicted positives, i.e. given by $\frac{TP}{TP+FP}$.
- *Recall*, also known as *sensitivity*, is the true positive rate, i.e. given by $\frac{TP}{TP+FN}$.
- *Specificity* is the true negative rate, i.e. given by $\frac{TN}{TN+FP}$.

In general, there is a trade-off between precision and recall. By adjusting probability thresholds, we can increase precision at the cost of decreasing recall, and vice versa. *F1 score* is a popular metric used to balance this trade-off between precision and recall, given by $\frac{TP}{TP+\frac{1}{2}(FP+FN)}$.

Similarly, there is a trade-off between sensitivity and specificity. Again, by adjusting probability thresholds, we can increase sensitivity at the cost of decreasing specificity, and vice versa. Cohen's *kappa score* (Feuerman & Miller, 2005) is a popular metric used to balance this trade-off between sensitivity and specificity. This score is given by the formula $\frac{2(TP \times TN - FN \times FP)}{(TP+FP) \times (FP+TN) + (TP+FN) \times (FN+TN)}$.

*Experiment results.* Our experiment results are reported in Table 2. Among all evaluated methods, we achieved the highest F1 score of 0.463, and the highest kappa score of 0.448. Note that our method achieves the highest recall/sensitivity of 0.508, which is a significant margin above the second highest value 0.488. For precision, our method has the value of 0.434, which is only marginally second to that of normalized entropy, 0.438. For specificity, all methods perform well, with specificity values at least 0.969. Our method has specificity 0.985, which is only marginally second to that of normalized entropy, 0.991.

Note that although normalized entropy (used in entropy maximization methods) has the highest precision and specificity (with our method coming a close second), its true positive rate (i.e. recall or sensitivity) is only 0.306, which is significantly lower than 0.508 achieved by our method. This means that our method is able to identify 20.2% more NC instances than normalized entropy.

### 5.3 Experiments on Web Image Classification

*Baselines.* We used 9 baselines in total, where 2 of the baselines, DSOS (Albert et al., 2022) and Jo-SRC (Yao et al., 2021), were already introduced in Sect. 2.1. The rest of the baselines are described as follows:

- *AFM* (Peng et al., 2020) introduces a training block that suppresses mislabeled data via grouping and self-attention.

- *CleanNet* (Lee et al., 2018) detects instances with label noise and assigns weights accordingly.
- *DivideMix* (Li et al., 2020) uses sample loss to partition the training data into a clean set and a noisy set. Then, two networks are trained jointly based on each network's data partition.
- *Joint optimization* (Tanaka et al., 2018) tackles label noise by alternately updating network parameters and labels during training.
- *MetaCleaner* (Zhang et al., 2019) uses a noisy weighting module to estimate weights for each instance and uses a clean hallucinating module to learn from weighted representations.
- *MoPro* (Li et al., 2021) identifies clean, noisy and OOD instances and assigns pseudo-labels accordingly. "Momentum prototypes" are then computed, after which both cross-entropy loss and contrastive loss are jointly used to train the model with the newly assigned pseudo-labels and momentum prototypes.
- *SMP* (Han et al., 2019) is an iterative self-training framework that measures data complexity and classifies data into several class prototypes. Models are trained on prototypes with the least complexity, which are assumed less likely to be noisy.

*Experiment Set-up.* Across all experiments on the Clothing1M, Food101/Food101N and mini WebVision 1.0 datasets, we used the same ResNet-50 architecture (He et al., 2016). For Clothing1M and Food101/Food101N, we initialize the ResNet-50 using weights pretrained on ImageNet. For mini WebVision 1.0, we follow the same set-up in our baselines (Li et al., 2020, 2021; Albert et al., 2022), and used the default random weight initialization.[7]

For the Clothing1M dataset, we first trained on the combined set of 1 million noisy instances and 50k clean instances, then fine-tuned on the 50k clean set.

For the Food101/Food101N datasets, we followed the popular experiment set-up, where the models are first trained on the combined set of 306k noisy instances and 53k clean instances of Food101N, then fine-tuned on the 53k clean instances of Food101N. Evaluation is then done on the Food101 test set.

For the mini WebVision 1.0 dataset, there is no clean set provided. To identify clean instances, we followed the common set-up in MoPro (Li et al., 2021) and DivideMix (Li et al., 2020), and let $X_{clean}$ vary over the epochs, where at the beginning of each epoch, we initialized $X_{clean}$ as the empty sequence, then computed $X_{clean}$ as follows: Using the weights from the previous epoch, an image is inserted into $X_{clean}$ if and only if the prediction value corresponding to its given label exceeds 0.5. For the first epoch, since there is no previous epoch, we instead trained a model with cross-entropy loss over 10 epochs, and used the trained model (trained over these 10 epochs) to identify $X_{clean}$.

---

[7] We used the PyTorch implementation of ResNet-50 with no pre-trained weights; see https://pytorch.org/vision/0.8/_modules/torchvision/models/resnet.html for default settings.

Note that full experimental set-up details (for all methods, across all datasets) are provided in Appendix C.2. In particular, we trained all models until convergence.

In stage two of our GENKL framework, across all datasets, the set $Q$ of prediction vectors is obtained by averaging over multiple models trained on $\mathcal{D}_{pre}$.[8] For subsequent iterations, the set $Q$ of prediction vectors is obtained by averaging from the models in previous iterations. Throughout, for the Clothing1M dataset and the Food101/Food101N datasets, we used SGD with initial learning rate 0.001, and Nesterov momentum 0.9, while for the mini WebVision 1.0 dataset, we used SGD with initial learning rate 0.01. For Clothing1M, we used $\omega_1 = 1, \omega_2 = 32, \omega_3 = 1$. For Food101/Food101N, we used $\omega_1 = 20, \omega_2 = 100, \omega_3 = 1$. For mini WebVision 1.0, we used $\omega_1 = 10, \omega_2 = 32, \omega_3 = 4$. For more training details (e.g. learning schedule), see Appendix C.

*Experiment results.* In comparison to all baselines, our proposed GENKL consistently achieved the highest test accuracies: 81.34% for Clothing1M, 85.73% for Food101/Food 101N and top-1/top-5 accuracies 78.99%/92.54% for mini WebVision 1.0. See Tables 3, 4 and 5 for full experiment results.

For Clothing1M, the next best replicable baseline (i.e. with publicly available code) after GENKL is DivideMix, which is on average 1.86% lower (79.48% versus our 81.34%). For Food101/Food101N, although Jo-SRC (Yao et al., 2021) and AFM (Peng et al., 2020) have test accuracies closest to ours, it should be noted that these two methods are the two lowest-performing baselines when evaluated on Clothing1M. As for mini WebVision 1.0, we outperformed the second best method (DivideMix) by a significant margin of 0.71% for top-1 accuracy (78.99% versus 78.28%), and a margin of 0.16% for top-5 accuracy (92.54% versus 92.38%).

## 5.4 Ablation Study

In Table 6, we evaluate the performances of GENKL on the Clothing1M dataset when individual components are removed.

*Averaging prediction vectors.* When the set $Q$ of prediction vectors was obtained from one single model (instead of being obtained by averaging from multiple models), we had

---

**Table 3** Averaged best test accuracies and standard deviations (over 5 trials) of different methods on the Clothing1M dataset

| # | Method | Accuracy (%) |
| --- | --- | --- |
| 1 | Jo-SRC (Yao et al., 2021) | $77.48 \pm 0.40$ |
| 2 | AFM (Peng et al., 2020) | $78.12 \pm 0.18$ |
| 3 | DSOS (Albert et al., 2022) | $78.33 \pm 0.25$ |
| 4 | Cross-entropy | $78.44 \pm 0.29$ |
| 5 | Joint Optimization (Tanaka et al., 2018) | $78.69 \pm 0.38$ |
| 6 | DivideMix (Li et al., 2020) | $79.48 \pm 0.37$ |
| 7 | MetaCleaner (Zhang et al., 2019) | 80.78* |
| 8 | SMP (Han et al., 2019) | 81.15* |
| 9 | GENKL | $\mathbf{81.34 \pm 0.18}$ |

We re-implemented our baselines wherever possible, using a common experiment set-up. For the two methods (#7, #8) that do not have publicly available code, we report the accuracies (marked with *) as indicated in the respective papers

**Table 4** Averaged best test accuracies and standard deviations (over 5 trials) of different methods on the Food101/Food101N datasets

| # | Method | Accuracy (%) |
| --- | --- | --- |
| 1 | Cross-entropy | $82.35 \pm 0.28$ |
| 2 | DivideMix (Li et al., 2020) | $83.29 \pm 0.08$ |
| 3 | DSOS (Albert et al., 2022) | $84.42 \pm 0.18$ |
| 4 | Joint Optimization (Tanaka et al., 2018) | $84.56 \pm 0.06$ |
| 5 | MetaCleaner (Zhang et al., 2019) | 85.05* |
| 6 | SMP (Han et al., 2019) | 85.11* |
| 7 | Jo-SRC (Yao et al., 2021) | $85.13 \pm 0.12$ |
| 8 | AFM (Peng et al., 2020) | $85.68 \pm 0.23$ |
| 9 | GENKL | $\mathbf{85.73 \pm 0.19}$ |

We re-implemented our baselines wherever possible, using a common experiment set-up. For the two methods (#5, #6) that do not have publicly available code, we report the accuracies (marked with *) as indicated in the respective papers

**Table 5** Averaged best test accuracies and standard deviations (over 5 trials) of different methods on the mini WebVision 1.0 dataset

| # | Method | Top-1 (%) | Top-5 (%) |
| --- | --- | --- | --- |
| 1 | Jo-SRC (Yao et al., 2021) | $66.32 \pm 1.35$ | $77.98 \pm 1.56$ |
| 2 | AFM (Peng et al., 2020) | $72.42 \pm 0.40$ | $90.54 \pm 0.26$ |
| 3 | Mopro (Li et al., 2021) | $73.35 \pm 0.44$ | $90.88 \pm 0.49$ |
| 4 | Cross-entropy | $73.96 \pm 0.37$ | $91.65 \pm 0.12$ |
| 5 | DSOS (Albert et al., 2022) | $76.78 \pm 0.84$ | $91.50 \pm 0.22$ |
| 6 | DivideMix (Li et al., 2020) | $78.28 \pm 0.37$ | $92.38 \pm 0.31$ |
| 7 | GENKL | $\mathbf{78.99 \pm 0.36}$ | $\mathbf{92.54 \pm 0.25}$ |

We re-implemented all baselines using a common experiment set-up

---

[8] For Clothing1M, $\mathcal{D}_{pre}$ is the 50k clean training set. For Food101/Food101N, we used the entire training set (with both noisy and clean instances) as our $\mathcal{D}_{pre}$. For mini WebVision 1.0, we similarly used the entire training set as our $\mathcal{D}_{pre}$.

**Table 6** Ablation study results on Clothing1M

| GENKL | Accuracy (%) |
|---|---|
| Complete framework | **81.34 ± 0.18** |
| w/o averaging prediction vectors | 81.31 ± 0.11 |
| w/o stratified sampling | 81.25 ± 0.17 |
| Eith double-hot vector normalization* | 81.18 ± 0.09 |
| w/o iterative relabeling | 80.94 ± 0.09 |

The averaged best test accuracies and standard deviations (over 5 trials) are reported for four experiments, each of which is conducted with one component removed from the complete GENKL framework. For the removal of the third component (marked with *), we are removing the non-stochasticity of double-hot vectors used in the relabeling process; effectively, for this particular experiment, we are *not* dropping the assumption that label vectors should be stochastic a test accuracy of 81.31%, which is a minor drop of 0.03% from the complete GENKL framework.

*Stratified sampling.* Our test accuracy dropped by 0.09% when vanilla random sampling was used in place of stratified sampling. This shows that our framework is not too affected by the imbalance between clean and non-clean (noisy and NC) instances.

*Double-hot vector normalization.* Recall that we used double-hot vectors for our iterative relabeling process in GENKL. By definition, these double-hot vectors are *not* stochastic vectors. We chose not to normalize these double-hot vectors because we could achieve higher test accuracies; cf. Section 3.6. For this part of our ablation study, we evaluated the effect of normalizing the double-hot vectors $\bar{q}^i$. By (4), the new $j$-th entry after normalization is $\frac{\bar{q}^i_j}{\lambda + \lambda^i}$. With this normalization, the resulting average test accuracy is 81.18%, which is a slight drop of 0.16% when compared to the original GENKL framework without normalization. In particular, this shows that the updated label vectors (via our relabeling process) need not be stochastic to achieve better performance.

*Iterative training.* Recall that GENKL is an iterative framework.[9] To understand the effect of iterations, we report in Table 6 the test accuracy when the training stops after the first iteration. Among all the components evaluated, the removal of this component has the most significant impact, yielding a 0.40% drop in test accuracy.

### 5.5 Sensitivity Analysis

Recall from Sect. 3.5 that our GENKL framework has seven hyperparameters $\alpha$, $\beta$, $|P|$, $\sigma$, $\omega_1$, $\omega_2$ and $\omega_3$. The first two hyperparameters ($\alpha$ and $\beta$) come from our $(\alpha, \beta)$-generalized KL divergence, $\mathcal{D}_{KL}^{\alpha,\beta}(p\|q)$. The next two hyperparameters ($|P|$ and $\sigma$) are used in stage one (identification of NC

**Table 7** Averaged best test accuracies and standard deviations (over 5 trials) on the Clothing1M dataset for our framework GENKL, over different values for $\alpha$: 0.90, 1.05 and 1.20

| $\alpha$ | 0.90 | 1.05 | 1.20 |
|---|---|---|---|
| Accuracy (%) | 81.20 ± 0.15 | **81.34 ± 0.18** | 81.19 ± 0.27 |

All other hyperparameter values are kept invariant

**Table 8** Averaged best test accuracies and standard deviations (over 5 trials) on the Clothing1M dataset for our framework GENKL, over different values for $\beta$: 0.02, 0.03 and 0.04

| $\beta$ | 0.02 | 0.03 | 0.04 |
|---|---|---|---|
| Accuracy (%) | 81.14 ± 0.23 | **81.34 ± 0.18** | 81.12 ± 0.07 |

All other hyperparameter values are kept invariant instances) of our GENKL framework to generate a set $P$ of uniform-like vectors with associated standard deviation $\sigma$. The last three hyperparameters ($\omega_1$, $\omega_2$ and $\omega_3$) are the weight factors for the respective three loss terms, (5), (6) and (7), in our loss function (8).

In Tables 7, 8, 9, 10, 11 and 12, we analyze the sensitivity of the performance of GENKL on the Clothing1M dataset, with respect to the values of the hyperparameters $\alpha$, $\beta$, $|P|$, $\sigma$, $\omega_2$, as well as with respect to the choice of the loss function for training. In particular, for our GENKL experiments on Clothing1M, we fixed $\omega_1 = 1$ and $\omega_3 = 1$ for simplicity. Hence, for our sensitivity analysis, we focused on $\omega_2$, which is the weight factor of the loss term (6) computed on non-NC instances.

*Sensitivity of hyperparameter $\alpha$.* Recall that $\alpha$ is the weight for the negative entropy term $-\alpha H(p)$ in our $(\alpha, \beta)$-generalized KL divergence $\mathcal{D}_{KL}^{\alpha,\beta}(p\|q)$; see (2). An instance is identified as an NC instance if $\mathcal{D}_{KL}^{\alpha,\beta}(p\|q) \geq 0$. Hence, as the value of $\alpha$ increases (note that $\alpha > 0$), the number of identified NC instances would decrease. We used $\alpha = 1.05$ in our GENKL experiments on the Clothing1M dataset. To see the effect of the value of $\alpha$, we also report the performance of GENKL for $\alpha = 0.90$ and $\alpha = 1.20$, while keeping other hyperparameter values the same; see Table 7. Our analysis demonstrates that the performance of GENKL is sensitive to the value of $\alpha$ (and hence the number of identified NC instances), where a deviation of $\pm 0.15$ in the value of $\alpha$ (from the optimal value $\alpha = 1.05$) resulted in an approximate 0.15% drop in accuracy.

*Sensitivity of hyperparameter $\beta$.* Recall that $\beta$ appears in the positive term $H(^\beta p, q)$ in our $(\alpha, \beta)$-generalized KL divergence $\mathcal{D}_{KL}^{\alpha,\beta}(p\|q)$. The larger the value of $\beta$, the larger this positive term $H(^\beta p, q)$ will be, and hence, more NC instances would be identified. Intuitively, $\beta$ controls the threshold for when an entry of the prediction vector is considered dominant, where if $\beta$ is too large, then almost all entries are considered dominant. We used $\beta = 0.03$ in our GENKL experiments on the Clothing1M dataset. Since Clothing1M

**Table 9** Averaged best test accuracies and standard deviations (over 5 trials) on the Clothing1M dataset for our framework GENKL, over different choices of the loss function: cross-entropy, MSE, MAE and KL loss

| Loss function | Cross-entropy | MSE | MAE | KL |
|---|---|---|---|---|
| Accuracy (%) | **81.34 ± 0.18** | 80.69 ± 0.15 | 80.16 ± 0.13 | 80.11 ± 0.32 |

All other hyperparameter values are kept invariant

**Table 10** Averaged best test accuracies and standard deviations (over 5 trials) on the Clothing1M dataset for our framework GENKL, over different values for $|P|$: 1, 10 and 20

| $|P|$ | 1 | 10 | 20 |
|---|---|---|---|
| Accuracy (%) | 80.26 ± 0.16 | 81.21 ± 0.18 | **81.34 ± 0.18** |

All other hyperparameter values are kept invariant

**Table 11** Averaged best test accuracies and standard deviations (over 5 trials) on the Clothing1M dataset for our framework GENKL, over different values for $\sigma$: 0.01, 0.05 and 0.10

| $\sigma$ | 0.01 | 0.05 | 0.10 |
|---|---|---|---|
| Accuracy (%) | 81.14 ± 0.27 | **81.34 ± 0.18** | 81.27 ± 0.08 |

All other hyperparameter values are kept invariant

**Table 12** Averaged best test accuracies and standard deviations (over 5 trials) on the Clothing1M dataset for our framework GENKL, over different values for $\omega_2$: 1, 32 and 64

| $\omega_2$ | 1 | 32 | 64 |
|---|---|---|---|
| Accuracy (%) | 80.26 ± 0.18 | **81.34 ± 0.18** | 81.32 ± 0.21 |

All other hyperparameter values are kept invariant has $k = 14$ classes, it means that our value $\beta = 0.03$ is slightly less than half of $\frac{1}{k} \approx 0.07143$. To see the effect of the value of $\beta$, we also report the performance of GENKL for $\beta = 0.02$ and $\beta = 0.04$, while keeping other hyperparameter values the same; see Table 8. Our analysis demonstrates that the performance of GENKL is sensitive to the value of $\beta$ (and hence the number of identified NC instances), where a deviation of $\pm 0.01$ in the value of $\beta$ (from the optimal value $\beta = 0.03$) resulted in an approximate 0.2% drop in accuracy.

*Sensitivity of hyperparameter $|P|$.* Recall that $P$ is the set of uniform-like vectors, such that each $p \in P$ is used to compute $\mathcal{D}_{\mathrm{KL}}^{\alpha, \beta}(p \| q)$ for the identification of NC instances. As we vary $p$ over all vectors in $P$, we take the union of all identified NC instances. Hence, as the value of $|P|$ increases (note that $|P| \geq 1$), the number of identified NC instances is expected to increase. We used $|P| = 20$ in our GENKL experiments on the Clothing1M dataset. To see the effect of the value of $|P|$, we also report the performance of GENKL for $|P| = 1$ and $|P| = 10$, while keeping other hyperparameter values the same; see Table 10.

*Sensitivity of hyperparameter $\sigma$.* Recall that $\sigma$ is the standard deviation of the normal distribution $\mathcal{N}(\frac{1}{k}, \sigma^2)$, used for sampling the value of each entry in a uniform-like vector $p \in P$ (before normalization). Hence, for a sufficiently large set $P$, as the value of $\sigma$ increases (note that $\sigma > 0$), the number of identified NC instances would tend to increase. We used $\sigma = 0.05$ in our GENKL experiments on the Clothing1M dataset. To see the effect of the value of $\sigma$, we also report the performance of GENKL for $\sigma = 0.01$ and $\sigma = 0.1$, while keeping other hyperparameter values the same; see

Table 11. Our analysis demonstrates that the performance of GENKL is sensitive to the value of $\sigma$ (and hence the number of identified NC instances), where the value $\sigma = 0.1$ may be a little too large, resulting in a slight 0.07% drop in accuracy (as compared to $\sigma = 0.05$).

*Sensitivity of hyperparameter $\omega_2$.* Recall that $\omega_2$ is the weight for the loss term $\mathcal{L}_{\mathrm{non\text{-}NC}}$ in (8). As the value of $\omega_2$ increases, the contribution of the non-NC instances to the overall loss would increase. We used $\omega_2 = 32$ in our GENKL experiments on the Clothing1M dataset. To see the effect of the value of $\omega_2$, we also report the performance of GENKL for $\omega_2 = 1$ and $\omega_2 = 64$, while keeping other hyperparameter values the same; see Table 12. Intuitively, when $\omega_2$ is sufficiently large (e.g. in the range of $\omega_2 = 32$ to $\omega_2 = 64$), the contribution of $\mathcal{L}_{\mathrm{non\text{-}NC}}$ to the overall loss has a regularization effect, thereby improving overall accuracy.

*Sensitivity of the choice of loss function.* Recall that we used cross-entropy loss as our loss function; see (5), (6), (7) and (8); cf. Section 3.6. To see the effect of the choice of loss function, we also report the performance of GENKL when the loss function is replaced by MSE, mean absolute error (MAE) and KL loss, respectively, while keeping all hyperparameter values the same; see Table 9. Our analysis demonstrates that our choice of cross-entropy loss is crucial for the outperformance of GENKL over the baselines.

## 6 Conclusion

We introduced the notion of non-conforming (NC) instances, which encompasses both ambiguous ID and OOD instances. Although there are numerous methods that tackle the problem of OOD instances, we are not aware of any method that explicitly tackles the problem of ambiguous ID instances, which are prevalent in web image datasets curated online. To tackle NC instances in a unified manner, we proposed a new generalized KL divergence, $\mathcal{D}_{\mathrm{KL}}^{\alpha, \beta}(p \| q)$, and an iterative training framework GENKL built upon this new generalized KL divergence. Moreover, we proved theoretical proper-

ties of $\mathcal{D}_{\text{KL}}^{\alpha,\beta}(p\|q)$. The key advantage of using $\mathcal{D}_{\text{KL}}^{\alpha,\beta}(p\|q)$ is that we can effectively identify more NC instances, including those whose predictions are not "almost" uniformly distributed, for which the usual approaches of entropy maximization and KL divergence minimization are unable to identify. We showed empirically that using $\mathcal{D}_{\text{KL}}^{\alpha,\beta}(p\|q)$ yields the best performance for NC instance identification. For our GENKL framework, we outperformed SOTA methods on real-world web image datasets: Clothing1M, Food101/Food101N and mini WebVision 1.0.

NC instances are unavoidable in web image datasets. Since the identification of NC instances is clearly a prerequisite step for tackling NC instances, we expect future work to further build upon the effectiveness of our new generalized KL divergence for NC instance identification.

**Acknowledgements** This work is supported by the National Research Foundation, Singapore under its AI Singapore Program (AISG Award No: AISG-RP-2019-015), and by Ministry of Education, Singapore, under its Tier 2 Research Fund (MOE2019-T2-1-178). The first author also thanks Shuqin Gao and Jingyi Xu for their insightful comments.

**Data Availability** In the interest of reproducibility, we have made our code available at https://github.com/codetopaper/GenKL. All experiments are conducted on publicly available datasets; see the references cited. For our experiments on the identification of NC instances, the list of the (manually verified) 200 NC instances in the Clothing1M dataset is also available at the same link.

## Appendix A: Proof of Theorem 2

*Lower Bound.*

Note that if $\beta = \frac{1}{k}$, then the indicator function $\mathbb{1}_{q_j \geq \frac{1}{k} - \beta}$ in (1) equals 1 and $\mathcal{D}_{\text{KL}}^{\alpha,\beta}(p\|q)$ becomes:

$$\mathcal{D}_{\text{KL}}^{\alpha,\beta}(p\|q) = \alpha \sum_{j=1}^{k} p_j \log p_j - \sum_{j=1}^{k} p_j \log q_j$$
$$\geq \alpha \sum_{j=1}^{k} p_j \log p_j - \sum_{j=1}^{k} p_j \log p_j$$

$$= -(1-\alpha)\sum_{j=1}^{k} p_j \log p_j = (1-\alpha)H(p),$$

where the inequality in the second line follows from Gibbs' inequality (see MacKay 2003, Section 2.6), with equality holding if and only if $p = q$. Note that the entropy $H(p)$ is bounded by

$$\log k \geq H(p) \geq 0. \quad (A1)$$

Equality holds for the first inequality in (A1) when $p$ is a uniform vector (i.e., $p = [\frac{1}{k}, \ldots, \frac{1}{k}]$), while equality holds for the second inequality in (A1) when $p$ is a one-hot vector.

Thus, if $\alpha > 1$, then $\mathcal{D}_{\text{KL}}^{\alpha,\beta}(p\|q) \geq (1-\alpha)\log k$, with equality holding if and only if $p$ is a uniform vector and $p = q$. If instead $0 < \alpha \leq 1$, then $\mathcal{D}_{\text{KL}}^{\alpha,\beta} \geq 0$, where equality holds if and only if $p$ is a one-hot vector and $p = q$.

Now, consider the case $0 \leq \beta < \frac{1}{k}$. Note that by (A1),

$$\alpha \sum_{j=1}^{k} p_j \log p_j = -\alpha H(p) \geq -\alpha \log k, \quad (A2)$$

with equality holding if and only if $p$ is a uniform vector. Note also that for all $j$, since $\log q_j \leq 0$ and $p_j \geq 0$, we have $-p_j \mathbb{1}_{q_j \geq \frac{1}{k} - \beta} \log q_j \geq 0$, hence

$$-\sum_{j=1}^{k} p_j \mathbb{1}_{q_j \geq \frac{1}{k} - \beta} \log q_j \geq 0. \quad (A3)$$

For equality to hold in (A3), every term $p_j \mathbb{1}_{q_j \geq \frac{1}{k} - \beta} \log q_j$ must be 0. For any $j$ such that $q_j \geq \frac{1}{k} - \beta$, this term equals $-p_j \log q_j$, which equals 0 if and only if $p_j = 0$ or $q_j = 1$. In particular, if $q_j = 1$ for some $j$, then $q$ is a one-hot vector. Thus, by (1), it follows from (A2) and (A3) that $\mathcal{D}_{\text{KL}}^{\alpha,\beta}(p\|q) \geq -\alpha \log k$, where equality holds if and only if $p$ is a uniform vector and $q$ is a one-hot vector.

*Upper bound.*

If $\beta = \frac{1}{k}$, then $\mathcal{D}_{\text{KL}}^{\alpha,\beta}(p\|q)$ diverges when there exists some $j$ such that $q_j = 0$ and $p_j \neq 0$.

Now consider $0 \leq \beta < \frac{1}{k}$. Note that

$$\alpha \sum_{j=1}^{k} p_j \log p_j - \sum_{j=1}^{k} p_j \mathbb{1}_{q_j \geq \frac{1}{k} - \beta} \log q_j$$
$$= \alpha \sum_{j=1}^{k} p_j \log p_j + \sum_{j=1}^{k} p_j \mathbb{1}_{q_j \geq \frac{1}{k} - \beta} \log\left(\frac{1}{q_j}\right)$$
$$(A4)$$

First, we note that the first term in (A4) is $\alpha \sum_{j=1}^{k} p_j \log p_j = -\alpha H(p) \leq 0$, which reaches maximum value 0 when $p$ is a one-hot vector. And note that the second term in (A4) becomes:

$$
\sum_{j=1}^{k} p_j \cdot \mathbb{1}_{q_j \geq \frac{1}{k} - \beta} \log\left(\frac{1}{q_j}\right)
$$
$$
\leq \left(\sum_{j=1}^{k} p_j\right) \max \left\{ \mathbb{1}_{q_j \geq \frac{1}{k} - \beta} \log\left(\frac{1}{q_j}\right) \mid 1 \leq j \leq k \right\}
$$
$$
\leq \max \left\{ \mathbb{1}_{q_j \geq \frac{1}{k} - \beta} \log\left(\frac{1}{q_j}\right) \mid 1 \leq j \leq k \right\}
$$
$$
\leq \log\left(\frac{1}{\frac{1}{k} - \beta}\right).
$$

Since $\log\left(\frac{1}{q_j}\right)$ is a strictly decreasing function in terms of $q_j$, the second term in (A4) reaches its maximum value $\log\left(\frac{1}{\frac{1}{k} - \beta}\right)$ when there exist indices $j_1, \ldots, j_\ell$ ($\ell \geq 1$) such that $q_{j_t} = \frac{1}{k} - \beta$ for all $1 \leq t \leq \ell$ and $\sum_{t=1}^{\ell} p_{j_t} = 1$. Therefore, (A4) reaches its maximum value $\log\left(\frac{1}{\frac{1}{k} - \beta}\right)$ when $p$ is a one-hot vector, say with a non-zero $j$-th entry, and $q_j = \frac{1}{k} - \beta$. □

## Appendix B: Proof of Theorem 3

We treat $\mathcal{D}_{\mathrm{KL}}^{\alpha,\beta}(p\|q)$ as a real-valued function on $\mathbb{R}^{2k}$, and consider its Hessian matrix $H$, which is a symmetric $2 \times 2$ block matrix, whose constituent blocks are $k \times k$ matrices, given as follows:

$$
H = \begin{bmatrix} A & B \\ B^T & C \end{bmatrix}.
$$

Note that $A$, $B$, and $C$ are diagonal matrices:

$$
A = \begin{bmatrix} \frac{\alpha}{p_1} & 0 & 0 \\ 0 & \ddots & 0 \\ 0 & 0 & \frac{\alpha}{p_k} \end{bmatrix},
$$
$$
B = \begin{bmatrix} -\frac{1}{q_1} & 0 & 0 \\ 0 & \ddots & 0 \\ 0 & 0 & -\frac{1}{q_k} \end{bmatrix},
$$
$$
C = \begin{bmatrix} \frac{p_1}{q_1^2} & 0 & 0 \\ 0 & \ddots & 0 \\ 0 & 0 & \frac{p_k}{q_k^2} \end{bmatrix}.
$$

Recall that the Schur complement of $A$ is defined to be $H/A := C - B^T A^{-1} B$. We check that:

$$
H/A = \begin{bmatrix} (1 - \frac{1}{\alpha})\frac{p_1}{q_1^2} & 0 & 0 \\ 0 & \ddots & 0 \\ 0 & 0 & (1 - \frac{1}{\alpha})\frac{p_k}{q_k^2} \end{bmatrix}.
$$

If $H$ is positive semi-definite, then $D_{\mathrm{KL}}^{\alpha,\beta}(p\|q)$ is convex. It follows from Smith (1992) that if $A$ is positive definite, then $H$ is positive semi-definite if and only if the Schur complement $H/A$ is positive semi-definite. Thus, to prove $D_{\mathrm{KL}}^{\alpha,\beta}(p\|q)$ is convex on $\mathbb{R}_u^\beta$, it suffices to show that $A$ is positive definite and show that the Schur complement $H/A$ is positive semi-definite.

If a symmetric matrix is strictly row diagonally dominant and has strictly positive diagonal entries, then it is positive definite. Given that $\alpha \geq 1$ and $A$ is symmetric, $A$ is positive definite. A symmetric diagonally dominant real matrix with non-negative diagonal entries is positive semi-definite. Given that $\alpha \geq 1$ and $p_i \geq 0$, $H/A$ is positive semi-definite. Thus, given $\alpha \geq 1$ and $(p, q) \in \mathbb{R}_u^\beta$, $H$ is positive semi-definite, which means $D_{\mathrm{KL}}^{\alpha,\beta}(p\|q)$ is piecewise convex. □

## Appendix C: Additional training details

This section contains more details (e.g. training hyperparameters) for the experiments in Sects. 5.2 and 5.3.

### C.1 Details for experiments in Sect. 5.2

For our baselines, we used the following hyperparameters:

- Jo-SRC (Yao et al., 2021): $\tau_{\mathrm{clean}} = 0.35$ and $\tau_{\mathrm{OOD}} = 0.5$.
- Delta divergence (Kittler & Zor, 2018): $\tau_A = 0.362$.
- DSOS (Albert et al., 2022): $\gamma = 0.37$ and $\delta = 0.0$.
- DC KL divergence (Ponti et al., 2017): $\tau_{\mathrm{DC}} = 0.43$.
- MSE: $\tau_{\mathrm{MSE}} = 0.00004$.
- Normalized entropy: $\tau_{\mathrm{Nor}} = 0.692$.
- KL divergence: $\tau_{\mathrm{KL}} = 2.12$.

For each validation set, we trained the model on its respective training set using cross-entropy loss with SGD, initial learning rate 0.01, Nesterov momentum 0.9, weight decay 0.001, and a batch size of 32, over 50 epochs. The learning rate was reduced by a factor of 10 whenever the validation loss did not drop after 4 consecutive epochs. The model from the epoch with the best validation accuracy is used to generate prediction vectors.

## C.2 Details for experiments in Sect. 5.3

For the Clothing1M dataset, we resized the images to 256×256, then randomly cropped them to size 224×224. Random horizontal flip was applied with probability 0.5.

Details of the hyperparameters used in the individual methods are given as follows:

- For our method, as part of the pre-training on the 50k clean set ($\mathcal{D}_{pre}$), we trained the model using cross-entropy loss with SGD, initial learning rate 0.01, Nesterov momentum 0.9, weight decay 0.001, and a batch size of 32, over 30 epochs. The learning rate was reduced by a factor of 10 whenever the validation loss did not drop after 4 consecutive epochs. To identify NC instances, we used the hyperparameters $\alpha = 1.05$, $\beta = 0.03$, $\sigma = 0.05$, and $|P| = 20$. During the main training stage (i.e. after the identification of NC instances), we used stratified sampling. This means that for each minibatch, half of the instances are sampled from the 50k clean set, while the remaining half of the instances are sampled from the 1 million noisy set. We also used mixup (Zhang et al., 2018) with hyperparameter 0.5. Weight decay is set at 0.001, and the learning rate was reduced by a factor of 10 whenever the validation loss did not drop after 3 consecutive epochs. We trained the model iteratively until convergence, with 20 epochs in each iteration. We then fine-tuned all methods on the 50k clean set over 25 epochs, using an Adam optimizer, with learning rate $5 \times 10^{-7}$ and weight decay 0.001.
- For Cross-entropy, we trained the model using cross-entropy loss with SGD, initial learning rate 0.001, Nesterov momentum 0.9, weight decay 0.001, and a batch size of 32, over 20 epochs. Learning rate was reduced by a factor of 10 whenever the validation loss did not drop after 3 consecutive epochs.
- For all other baselines that we re-implemented, we used the same hyperparameters as those reported in their respective papers.

For the Food101/Food101N datasets, in addition to the same application of resizing, random cropping and random horizontal flip, as described above for the Clothing1M dataset, we also applied random rotation within the range of ±30 degrees on the images. For our method and all re-implemented baselines, unless otherwise mentioned, we always used the following hyperparameters in the main training stage: We trained the model with SGD, using initial learning rate 0.001, Nesterov momentum 0.9, weight decay 0.005, and a batch size of 128, over 40 epochs. The learning rate was adjusted using cosine annealing. We fine-tuned on the 53k clean set over 40 epochs, using regular random sampling, cross-entropy loss and an Adam optimizer, with learning rate $10^{-10}$ and weight decay $10^{-4}$.

Exceptions to these hyperparameters are described as follows:

- For our method, during training on $\mathcal{D}_{pre}$ in the first iteration, we trained the model using cross-entropy loss with SGD, initial learning rate 0.001, Nesterov momentum 0.9, weight decay 0.005, and a batch size of 128, over 50 epochs. The learning rate is adjusted using cosine annealing. To identify NC instances, we used the hyperparameters $\alpha = 1.1$, $\beta = 0.008$, $\sigma = 0$, and $|P| = 1$.
- For Joint Optimization (Tanaka et al., 2018), in the first step, we trained the model with SGD, using initial learning rate 0.003, Nesterov momentum 0.9, weight decay 0.0001, and a batch size of 128, over 30 epochs. For the hyperparameters $\alpha$, $\beta$ specific to Joint Optimization, we used $\alpha = 0.7$ and $\beta = 0.4$. In the second step, we trained the model with SGD, using initial learning rate 0.001, Nesterov momentum 0.9, weight decay 0.0001, and a batch size of 128, over 40 epochs.
- For Jo-SRC (Yao et al., 2021) and AFM (Peng et al., 2020), we used the same hyperparameters as those reported in their respective papers or in the code the authors released.

For the mini WebVision 1.0 dataset, we resized the images to 320×320, then randomly cropped them to size 299×299. Random horizontal flip was applied with probability 0.5.

Details of the hyperparameters used in the individual methods are given as follows:

- For our method, we used the training set as our $\mathcal{D}_{pre}$ dataset for pre-training. As part of the pre-training on the whole noisy training set ($\mathcal{D}_{pre}$), we trained the model using cross-entropy loss with SGD, initial learning rate 0.01, momentum 0.9, weight decay 0.001, and a batch size of 32, over 100 epochs. The learning rate was reduced by a factor of 10 in epoch 50. To identify NC instances, we used the hyperparameters $\alpha = 0.9$, $\beta = 0.015$, $\sigma = 0.05$, and $|P| = 20$. During the main training stage (i.e. after the identification of NC instances), we followed the set-up in MoPro (Li et al., 2021) and DivideMix (Li et al., 2020) to select clean instances: If the entry of the prediction vector corresponding to the given label is above 0.5, then this instance is clean, i.e. in $\mathcal{D}_{clean}$. We used mixup (Zhang et al., 2018) with hyperparameter 3.0. Weight decay is set at 0.0001. Learning rate is adjusted using cosine annealing. We trained the model for 300 epochs. Subsequently, we used $X_{clean}$ from epoch 300 for fine-tuning over 50 epochs, using an Adam optimizer, with learning rate $5 \times 10^{-7}$ and weight decay 0.001.

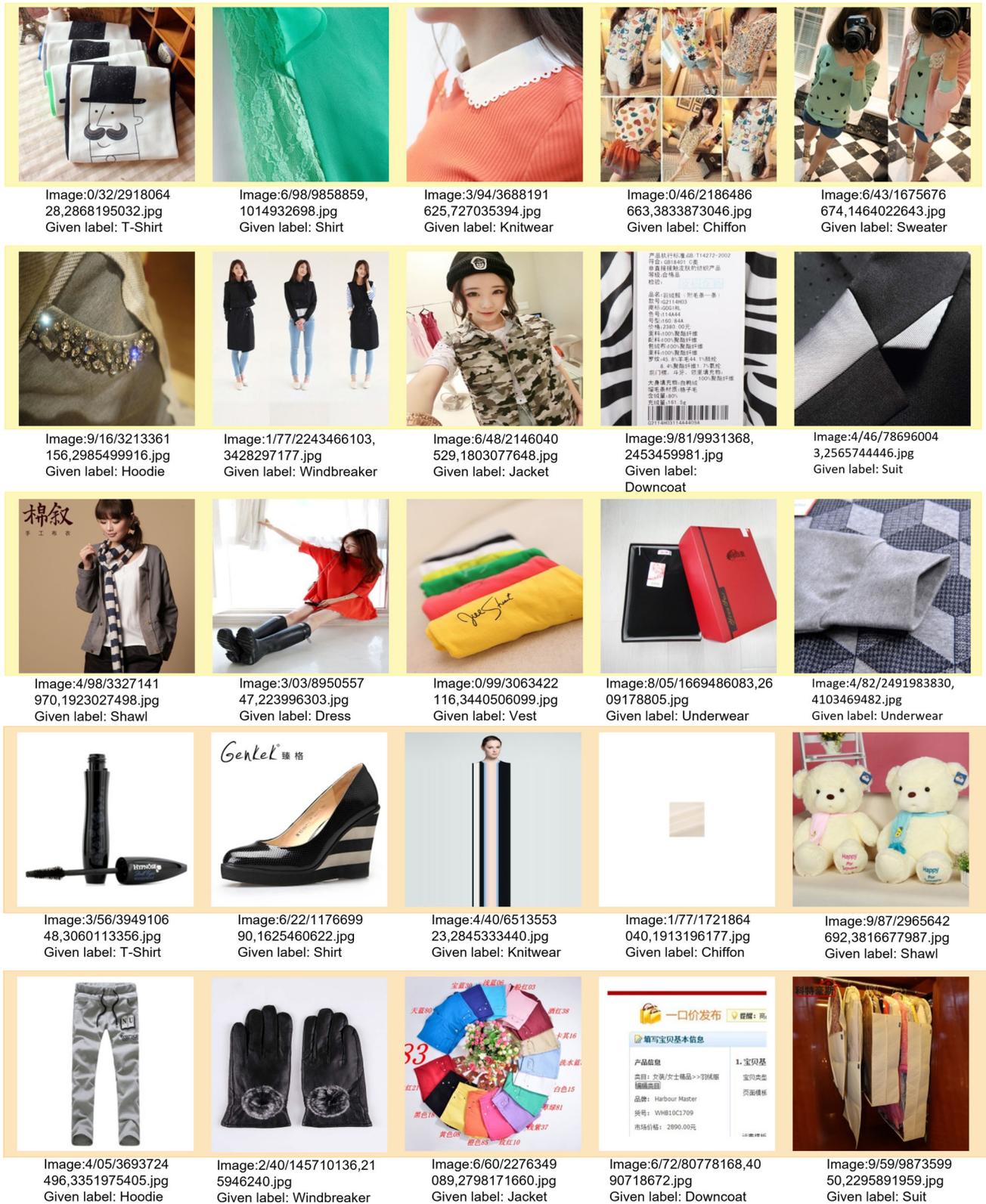

**Fig. 4** A depiction of some NC instances found in the Clothing1M dataset. The images in the last two rows depict OOD instances. The images in the first three rows depict Ambiguous ID (AID) instances. The 14 given labels in Clothing1M are: T-Shirt, Shirt, Knitwear, Chiffon, Sweater, Hoodie, Windbreaker, Jacket, Downcoat, Suit, Shawl, Dress, Vest, Underwear

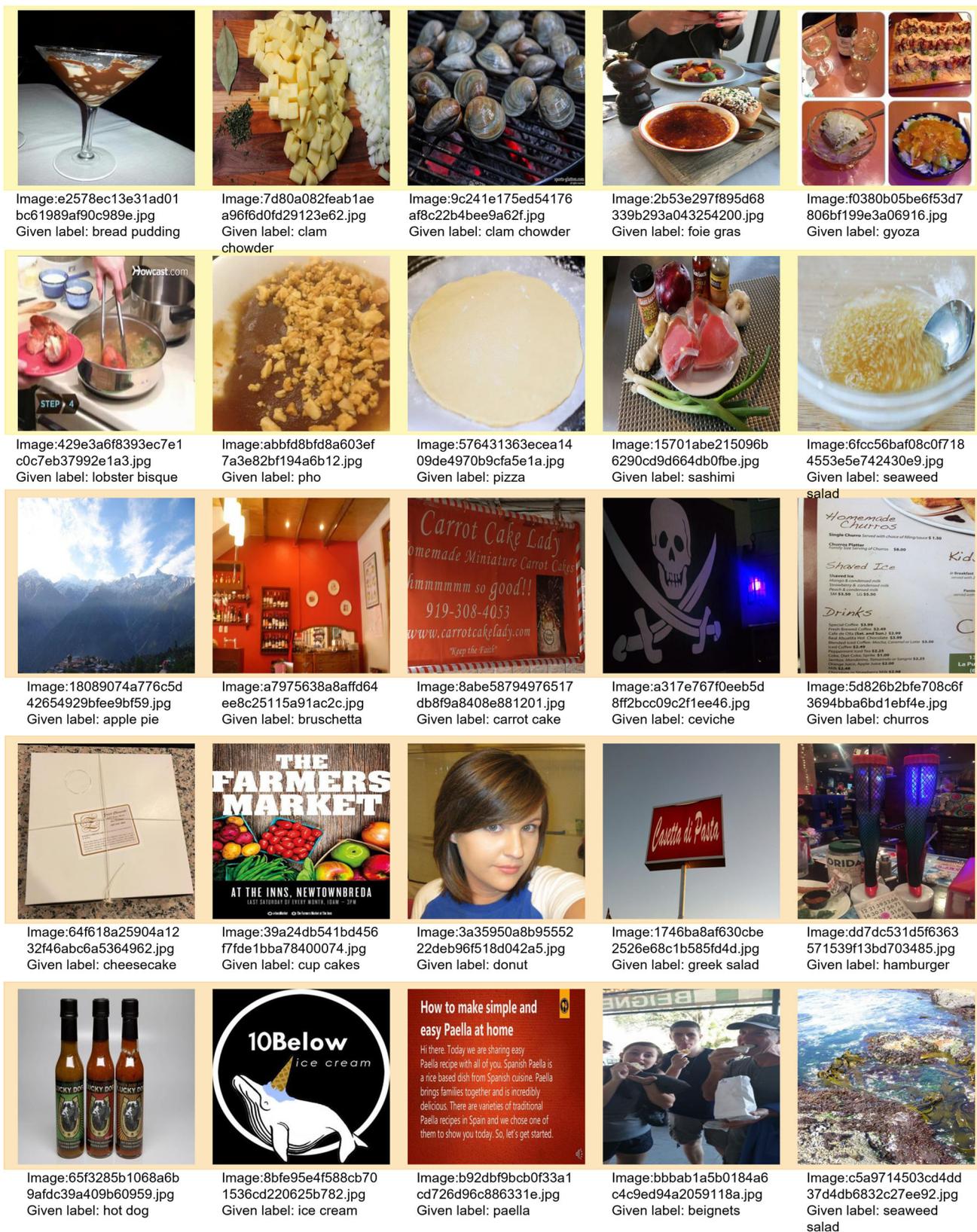

**Fig. 5** A depiction of some NC instances found in the Food101N dataset. The images in the last three rows depict OOD instances. The images in the first two rows depict Ambiguous ID (AID) instances

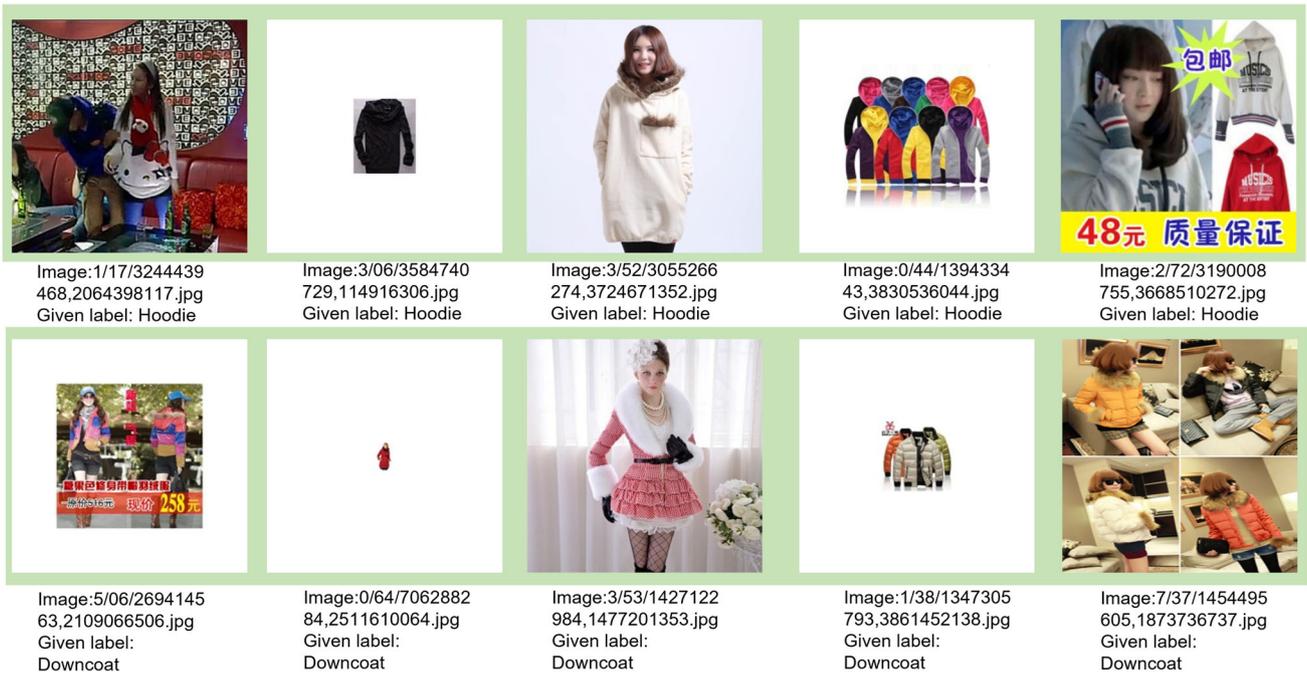

**Fig. 6** A depiction of some non-NC instances that are wrongly identified as NC instances via our $(\alpha, \beta)$-generalized KL divergence for classes Hoodie and Downcoat on the Clothing1M dataset

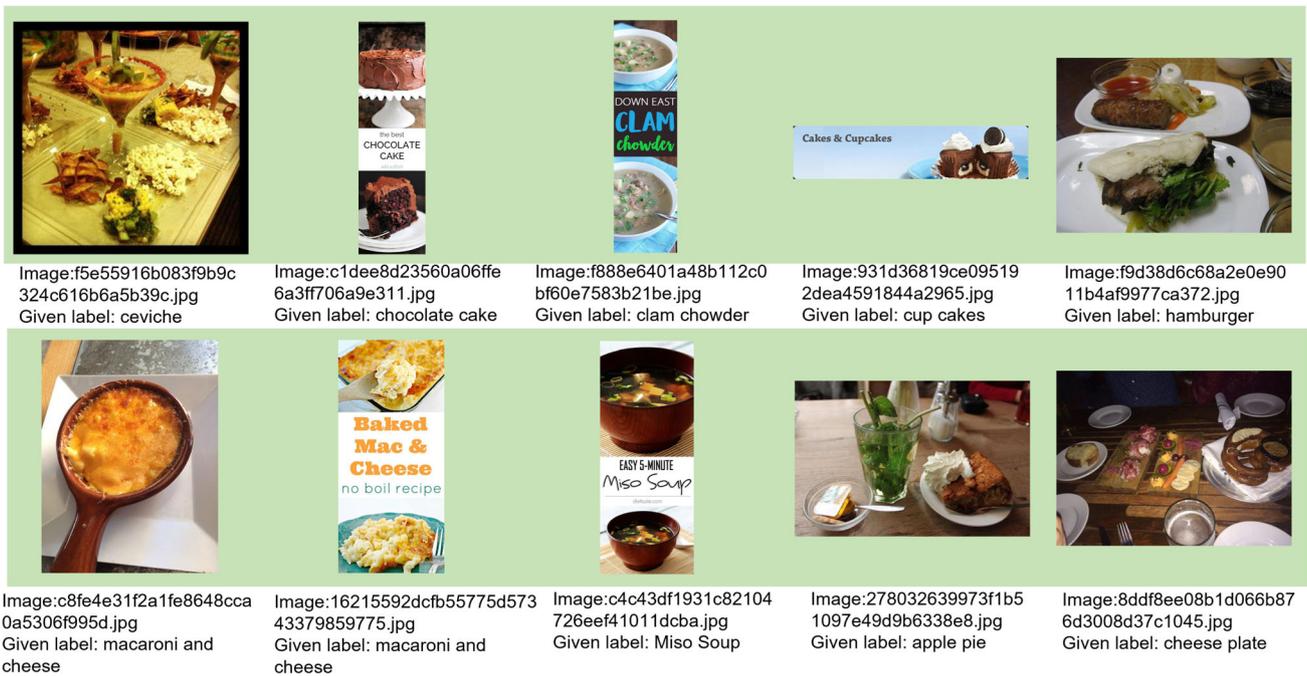

**Fig. 7** A depiction of some non-NC instances that are wrongly identified as NC instances via our $(\alpha, \beta)$-generalized KL divergence on the Food101N dataset

- For DivideMix (Li et al., 2020) and DSOS (Albert et al., 2022), which we re-implemented, we used the same hyperparameters as those reported in Li et al. (2020), Albert et al. (2022).
- For Cross-entropy, we trained the model using cross-entropy loss with SGD, initial learning rate 0.01, Nesterov momentum 0.9, weight decay 0.001, and a batch size of 32, over 100 epochs. Learning rate was reduced by a factor of 10 at epoch 50.
- MoPro (Li et al., 2021) has its implementation on the Webvision 1.0 (Li et al., 2017) dataset, and we used its respective hyperparameters for mini WebVision 1.0 (Jiang et al., 2018).
- For AFM (Peng et al., 2020), we used the same optimizer, learning rate, momentum, weight decay, batch size, epoch numbers and scheduler as those used for the Cross-entropy baseline. We used hyperparameter values $\ell_{afm} = 0.5$, $\ell_{org} = 1.5$.
- For JoSRC (Yao et al., 2021), we used the same optimizer, learning rate, momentum, weight decay, batch size, epoch numbers and scheduler as those used for the Cross-entropy baseline. The number of warm-up epochs is 10. We used the values $\tau_{clean} = 0.9$, $\alpha = 0.2$. The hyperparameter *eps* has the same value as that used for Clothing1M, which is 0.7.

## Appendix D: Demonstration of NC instances

This section demonstrates some of the NC instances found in the Clothing1M dataset and the Food101N dataset using our $(\alpha, \beta)$-generalized KL divergence, given in Figs. 4 and 5, respectively. In Figs. 6 and 7, we show some non-NC instances that have been wrongly identified as NC instances in the Clothing1M dataset and the Food101N dataset, respectively. We posit that some possible reasons for the wrong identification could be: (i) the background of the image is too noisy, (ii) the image content takes up a small portion of the image, (iii) the image is inherently hard to classify, (iv) there are multiple objects of interest in the image, (v) the object of interest does not take up the center portion of the image (note that during training, the image is first resized then randomly cropped; this could potentially generate a training image without the object of interest), and (vi) bad lighting.